\PassOptionsToPackage{table}{xcolor}
\documentclass[10pt,twocolumn,letterpaper]{article}

\usepackage[pagenumbers]{cvpr}      

\usepackage{graphicx}
\usepackage{booktabs}
\usepackage{multirow}            
\usepackage[table]{xcolor}       
\usepackage{colortbl}
\usepackage{float}               
\usepackage{dblfloatfix}         
\usepackage{currfile}            
\usepackage{caption}             

\AtBeginDocument{%
  \providecommand{\expImage}{}%
  \renewcommand{\expImage}[2]{%
    \IfFileExists{#2.pdf}{\includegraphics[width=#1,height=0.38\textheight,keepaspectratio]{#2.pdf}}{%
    \IfFileExists{#2.png}{\includegraphics[width=#1,height=0.38\textheight,keepaspectratio]{#2.png}}{%
    \IfFileExists{#2.jpg}{\includegraphics[width=#1,height=0.38\textheight,keepaspectratio]{#2.jpg}}{%
      \fbox{\parbox{\dimexpr#1-2\fboxsep-2\fboxrule\relax}{\centering\vspace{2em}%
        \texttt{\detokenize{#2}} (missing)\vspace{2em}}}%
    }}}%
  }%
}

\usepackage{placeins}
\usepackage{graphicx}
\usepackage{booktabs}
\usepackage{multirow}
\usepackage{flafter}

\definecolor{cvprblue}{rgb}{0.21,0.49,0.74}
\usepackage[pagebackref,breaklinks,colorlinks,allcolors=cvprblue]{hyperref}

\def\confName{CVPR}
\def\confYear{2027}

\title{
GroundedDream: Aligning Physical Dynamics in World Models for Policy Optimization
}

\author{Hui Liu\textsuperscript{1}\quad Fangqi Zhu\textsuperscript{1}\quad Junhao Li\textsuperscript{1}\quad Xiaoyi Pang\textsuperscript{1}\\
Huang Jiarui\textsuperscript{1}\quad Zhuo Han\textsuperscript{1}\quad Yue Lin\textsuperscript{1}\quad Ying Chen\textsuperscript{1}\\
Hongrui Yin\textsuperscript{2}\quad Xuanzhi Feng\textsuperscript{1}\quad Jie ZHANG\textsuperscript{1}\quad Song Guo\textsuperscript{1}\\
\textsuperscript{1}The Hong Kong University of Science and Technology\\
\textsuperscript{2}The Chinese University of Hong Kong, Shenzhen}
\hypersetup{pdftitle={GroundedDream: Aligning Physical Dynamics in World Models for Policy Optimization},pdfauthor={Hui Liu, Fangqi Zhu, Junhao Li, Xiaoyi Pang, Huang Jiarui, Zhuo Han, Yue Lin, Ying Chen, Hongrui Yin, Xuanzhi Feng, Jie ZHANG, Song Guo}}
\begin{document}
\maketitle
\begin{abstract}
Generative world models hold great promise for scalable robot policy evaluation and optimization in imagined environments. However, their pixel-level training objectives emphasize appearance, so rollouts often look realistic but mispredict the outcomes of actions, especially when errors unfold across multiple video chunks. We introduce GroundedDream, a post-training framework for video world models based on Physical Dynamics Alignment (PDA). Instead of comparing rollouts in pixel space, PDA learns a conditional critic in a predictive latent space that distinguishes the model's own multi-chunk rollouts from reference rollouts. Because the critic is differentiable, the world model can directly minimize the Wasserstein distance estimated by the critic, aligning all chunks of a rollout jointly while flow matching preserves visual fidelity. The aligned model then serves as an imagined environment for policy evaluation and model-based reinforcement learning (RL) of vision-language-action models (VLAs) and world action models (WAMs). Experiments in simulation and on four real-robot tasks show that GroundedDream (1) improves both the physical correctness and the visual quality of long-horizon predictions, (2) evaluates policies more accurately, and (3) yields better policies from model-based RL than other world models, improving real-robot success by 20.0 percentage points over the initial supervised fine-tuned (SFT) policy.

\end{abstract}

\section{Introduction}
\label{sec:introduction}

\begin{figure*}[!tb]
    \centering
    \includegraphics[width=\textwidth,trim=5bp 30bp 6bp 36bp,clip]{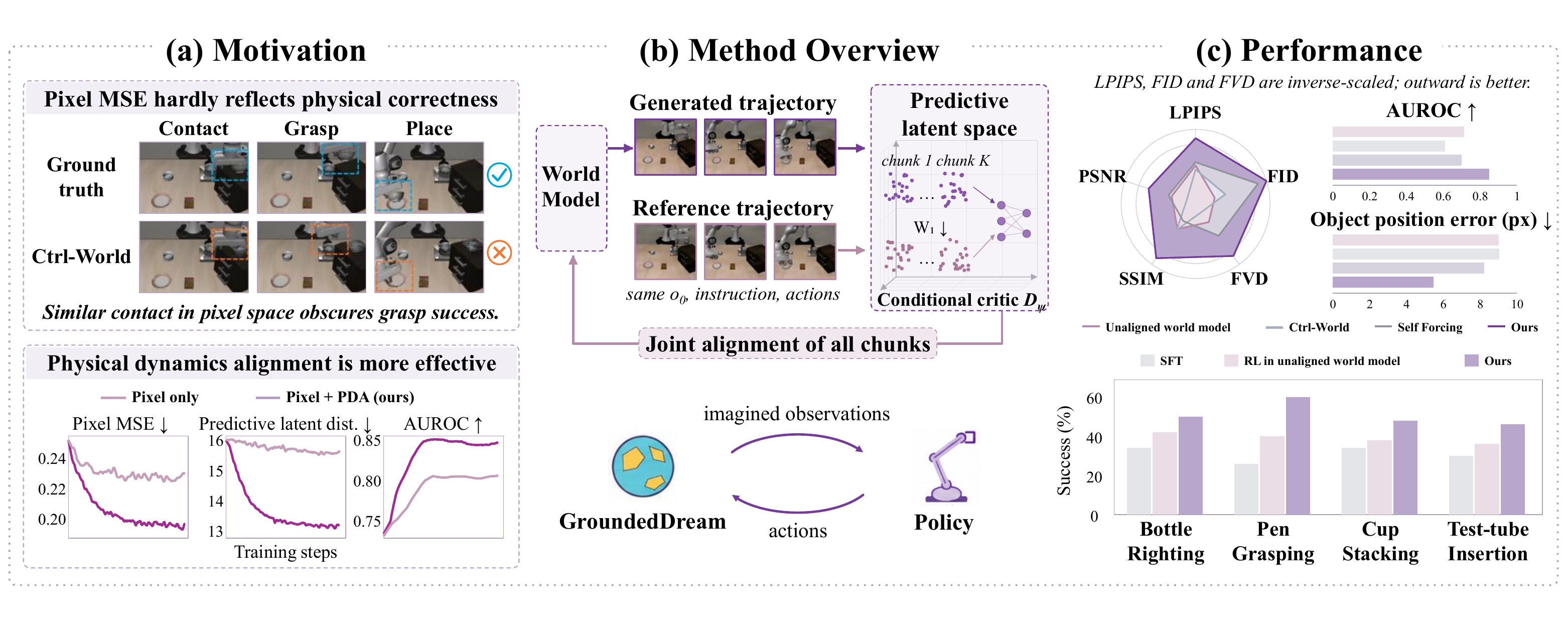}
    \caption{\textbf{Overview of GroundedDream.}
    (a) Motivation. In the top row, a Ctrl-World~\cite{guo2026ctrlworld} rollout looks similar to the ground truth but differs at contact, grasp, and placement, and such errors change the pixel loss only slightly. In the bottom row, adding Physical Dynamics Alignment (PDA) to pixel-level training brings rollouts much closer to reference rollouts in the predictive latent space and more often ranks successful action sequences above failed ones (AUROC), while pixel MSE also decreases.
    (b) Method. A conditional critic $D_\psi$ compares the world model's multi-chunk rollouts with reference rollouts that share the same initial observation, instruction, and actions in a predictive latent space, and the world model reduces the estimated Wasserstein distance $W_1$ for all chunks jointly. The aligned world model then serves as an imagined environment for policy evaluation and optimization.
    (c) Performance. GroundedDream improves visual quality, AUROC, and object position error over the Unaligned world model, Ctrl-World, and Self Forcing, and raises real-robot success on four tasks over the SFT policy and RL in the Unaligned world model.}
    \label{fig:intro}
\end{figure*}

World models learn how an environment evolves in response to actions, enabling an agent to anticipate the consequences of its behavior before acting in the real world~\cite{ha2018world,hafner2023dreamerv3,parkerholder2025genie3}. For robot learning, they offer a scalable alternative to physical trial and error, which is costly, slow, and unsafe~\cite{yang2023unisim,zhu2025irasim}. Driven by advances in video generation~\cite{brooks2024sora,agarwal2025cosmos,wan2025wan}, world models can now synthesize realistic, long-horizon rollouts of robot manipulation~\cite{guo2026ctrlworld,zhu2025wmpo}. Yet realism alone does not guarantee physical correctness; a convincing rollout may still mispredict whether a grasp holds or how an object moves after contact~\cite{motamed2026physicsiq,dong2026egovideo}. Consequently, success rates estimated in the world model can diverge from those measured on the real robot~\cite{wu2025rlvrworld}, and policies optimized within it may acquire behaviors that succeed in imagination but fail upon deployment~\cite{jiang2026wovr}.

This gap between realism and physical correctness traces back to how video world models are pretrained. Their denoising objectives~\cite{ho2020ddpm,lipman2023flow} scale well to large video corpora but are less effective at capturing physical dynamics. Scaling data or training narrows this gap only slowly. A recent large-scale study~\cite{dong2026egovideo} shows that a 100-fold increase in training data, up to 30,000 hours of egocentric video, yields limited gains in object interactions, attributing the bottleneck to how models are trained. We find in a diagnostic experiment (Fig.~\ref{fig:intro}a, bottom) that continued pixel-level training likewise lowers pixel error only slowly and barely reduces the distance to reference rollouts in a predictive latent space, while its ranking of successful above failed action sequences (AUROC) stops improving early. This limitation stems from the denoising objective itself, which weighs all parts of an observation equally and spends capacity on unpredictable visual details~\cite{assran2025vjepa2}, so a small contact region that often decides task success~\cite{zhu2025irasim} receives little weight (Fig.~\ref{fig:intro}a, top). Physical correctness also depends on the order of events, which often unfold across several chunks of a rollout and are hard to assess with a per-chunk loss. An object may, for instance, snap into the gripper before any contact is made~\cite{guo2026ctrlworld}, depicting a grasp that never occurred. What is needed after pretraining is therefore a training signal that distinguishes physically faithful rollouts from unfaithful ones, emphasizing how the scene changes rather than how it looks and assessing these changes jointly across a trajectory instead of chunk by chunk.

Existing post-training methods meet these two needs only in part. Distribution-matching methods such as Self Forcing~\cite{huang2025selfforcing,yin2025causvid} train on multi-chunk self-rollouts but align them with a teacher's video distribution, so their target remains the teacher's appearance rather than physical dynamics. RL-based methods optimize rewards such as perceptual similarity~\cite{wu2025rlvrworld} or inverse-dynamics action recovery~\cite{ye2025rlir}, but compute them on individual predictions or short clips, leaving errors that unfold across chunks unassessed.

In this work, we propose GroundedDream, a post-training framework for video world models based on Physical Dynamics Alignment (PDA). PDA learns this signal with a conditional critic that compares the model's own multi-chunk rollouts with reference rollouts sharing the same observation, instruction, and actions (Fig.~\ref{fig:intro}b). Consistent differences between them therefore indicate where the world model mispredicts the outcome of the actions. To focus on physical dynamics, the critic operates in the predictive latent space of a frozen V-JEPA~2 encoder~\cite{assran2025vjepa2}, which reflects how the scene evolves rather than how it looks. To assess these dynamics at the trajectory level, it scores several consecutive chunks jointly. The critic is trained to estimate the Wasserstein distance between generated and reference trajectory distributions~\cite{arjovsky2017wgan,chemseddine2025conditional}, and the world model is trained to reduce this distance together with a self-rollout flow-matching loss that keeps its predictions visually faithful. Unlike RL fine-tuning~\cite{black2024ddpo,liu2025flowgrpo}, which assigns a single scalar to each sampled rollout and compares groups of samples to estimate an update, the critic is differentiable, so its trajectory-level assessment yields a separate gradient for every chunk from a single rollout. The aligned model is then frozen as an imagined environment, in which policies are evaluated and improved by model-based RL from the outcomes of complete rollouts. This applies to both VLAs and WAMs, and for WAMs these outcomes update both the visual and the action components of the policy.

We evaluate GroundedDream on LIBERO~\cite{liu2023libero} and four real-robot tasks across three aspects: prediction, policy evaluation, and policy optimization. Its long-horizon predictions better distinguish successful from failed action sequences, nearly halving how often a failed sequence is ranked above a successful one (AUROC from 0.712 to 0.849), and reduce the error in predicted object positions by 39.4\%, while visual quality also improves (Fig.~\ref{fig:intro}a and c). Its success estimates are closer to actual success rates and better preserve the ranking of policy checkpoints. In model-based RL, policies optimized inside it outperform those optimized in other world models for a WAM and two VLAs, and on the real robot they improve success by 20.0 percentage points over the initial SFT policy and by 12.0 points over RL in the Unaligned world model (Fig.~\ref{fig:intro}c).

\section{Related Work}
\label{sec:related}

\paragraph{World models.}
Robotic world models incorporate heterogeneous data~\cite{yang2023unisim,jang2025dreamgen}, jointly model observations and actions~\cite{wu2024ivideogpt}, or adapt pretrained video models~\cite{rigter2024avid,du2024vlp} to build imagined environments for closed-loop policy interaction~\cite{guo2026ctrlworld,quevedo2026worldgym,gao2026dreamdojo,huang2026a2world}.
However, reconstruction or denoising objectives do not explicitly ensure accurate predictions of action consequences~\cite{wu2025rlvrworld,chen2026worldecho}.
Another line of work learns dynamics in latent space~\cite{hansen2024tdmpc2,zhou2025dinowm,assran2025vjepa2}, but typically does not directly generate image observations and thus cannot readily serve as an interaction environment for visual policies.
GroundedDream instead supervises dynamics in a predictive latent space while retaining image generation for closed-loop interaction with visual policies.

\paragraph{Predictive representations for physical plausibility.}
Self-supervised video representations capture physical dynamics more reliably than the internal features of video generators~\cite{zhang2025videorepa,assran2025vjepa2}, and have been used to improve the physical plausibility of video generation: as relational alignment targets during fine-tuning~\cite{zhang2025videorepa,xiong2026physalign}, as inference-time rewards~\cite{yuan2026physicsalignment}, and as feature teachers and reward critics in post-training~\cite{luo2026ldo,zhang2025mindv}.
These methods assess whether a generated video is physically plausible on its own, without reference to the actions that produced it.
GroundedDream instead aligns generated and reference trajectories under matched initial observations, instructions, and actions to constrain action-specific outcomes.

\paragraph{Post-training of world models.}
Existing post-training approaches differ in the source of supervision.
Distribution-matching methods use teacher-generated targets to improve autoregressive video generation~\cite{yin2025causvid,huang2025selfforcing,cui2025selfforcingpp,liu2025rolling}, whereas reward-based methods derive supervision from visual quality~\cite{wu2025rlvrworld}, human preferences~\cite{liu2025videoalign}, or inverse-dynamics action recovery~\cite{ye2025rlir}.
Adversarial objectives in frozen feature spaces have also been used to post-train image and video generators~\cite{sauer2024add,lin2025apt}, but they match appearance distributions without action conditioning.
GroundedDream instead aligns multi-chunk rollouts with recorded interaction trajectories rather than teacher-generated targets or predefined reward criteria.
An extended discussion is \mbox{provided in Appendix~A.}

\section{Physically Grounded World Model for Policy Evaluation and Optimization}
\label{sec:method}

\begin{figure*}[!tb]
    \centering
    \includegraphics[width=\textwidth]{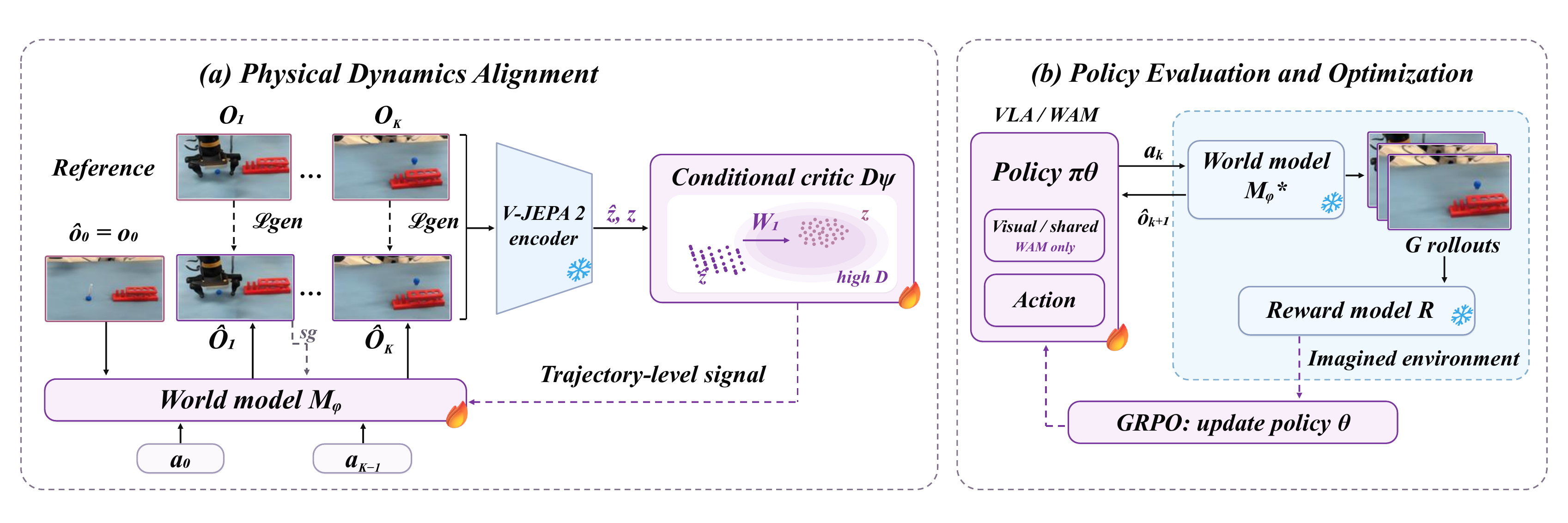}
    \caption{\textbf{Overview of GroundedDream.}
    \textbf{(a) Physical Dynamics Alignment.}
    A conditional critic aligns generated and reference $K$-chunk trajectories in the latent space of a frozen V-JEPA~2 encoder under matched initial observations, instructions, and actions.
    Chunk-wise flow matching supplements trajectory alignment; sg denotes stop-gradient between chunks.
    \textbf{(b) Policy evaluation and optimization.}
    The aligned world model and reward model remain frozen during closed-loop policy evaluation and optimization.
    Terminal rewards provide group-relative advantages; WAM action-likelihood gradients pass through the policy's visual and action components.
    Flames and snowflakes indicate trainable and frozen components, respectively.}
    \label{fig:method}
\end{figure*}

Our goal is to learn a world model that supports reliable policy evaluation and effective policy optimization. As shown in Fig.~\ref{fig:method}, GroundedDream first aligns the world model with reference interaction trajectories through PDA, then uses it for closed-loop evaluation and optimization of VLA policies and WAMs. Implementation details are \mbox{provided in Appendix~B.}

\subsection{Physical Dynamics-Aligned World Model}
\label{sec:pda}

We initialize the world model $M_\phi$, parameterized by $\phi$, from Cosmos~3 Edge~\cite{agarwal2026cosmos3} and post-train it with policy rollout data and a trajectory-level alignment objective. For a training trajectory containing $K$ action chunks, we use $g$ to denote the task instruction and $k\in\{0,\ldots,K-1\}$ to index chunks. Action chunk $a_k$ produces the subsequent RGB observation chunk $O_{k+1}$; thus, $a_{K-1}$ produces the terminal chunk $O_K$. The last frame of each observation chunk becomes the starting observation for the next chunk. Hats denote generated observations, trajectories, or \mbox{their representations.}

\paragraph{Policy rollout data and self-rollout training.}
Demonstrations predominantly contain successful executions, leaving limited coverage of states reached when a policy makes mistakes. A world model trained only on demonstrations may therefore become overly optimistic when the policy deviates from them. In addition, training conditions on reference observations, whereas closed-loop inference conditions on generated observations, allowing errors to accumulate across chunks. We address both mismatches by combining policy rollout data with self-rollout training, exposing the model to policy-induced deviations and its own imperfect predictions.

Specifically, we collect a policy rollout dataset $\mathcal D_\pi$ containing successful and failed executions and combine it with demonstration data $\mathcal D_{\mathrm{demo}}$, yielding $\mathcal D=\mathcal D_{\mathrm{demo}}\cup\mathcal D_\pi$. Policy rollouts cover policy-induced states, while demonstrations retain coverage of successful task execution. For a reference trajectory $\tau=(O_1,\ldots,O_K)$ sampled from $\mathcal D$, the model starts from the initial frame $\hat o_0=o_0$ and follows its $K$ recorded action chunks $a_{0:K-1}$. Each chunk is conditioned on the last frame $\hat o_k$ of the previous prediction, producing $\hat\tau=(\hat O_1,\ldots,\hat O_K)$. We apply auxiliary flow-matching supervision~\cite{lipman2023flow} to future observation latents:
\begin{equation}
\begin{aligned}
\mathcal L_{\mathrm{gen}}^{\mathrm{SR}}(\phi)
=
\mathbb E\Big[
\big\|
&v_\phi\!\left(
x_\sigma,\sigma\mid
\operatorname{sg}(\hat o_k),g,a_k
\right)\\
&-(\epsilon-Z_{k+1})
\big\|_2^2
\Big].
\end{aligned}
\label{eq:gen_sr}
\end{equation}
Here, $Z_{k+1}$ is the encoding of $O_{k+1}$ produced by the frozen variational autoencoder (VAE), $x_\sigma=(1-\sigma)Z_{k+1}+\sigma\epsilon$, $\epsilon\sim\mathcal N(0,I)$, and $\sigma$ denotes the noise fraction. The velocity target $\epsilon-Z_{k+1}$ follows the increasing-noise convention. The expectation covers trajectories, generated histories, chunk indices, and noise sampling; $\operatorname{sg}$ denotes stop-gradient and SR denotes self-rollout conditioning.

\paragraph{Conditional trajectory alignment in predictive latent space.}
Chunk-wise generative supervision does not directly assess whether action-induced motion and interactions remain consistent throughout a trajectory. We therefore introduce conditional trajectory alignment to compare generated and reference spatiotemporal representations under the same actions. We match distributions rather than regress paired features because sampled rollouts are stochastic, and a per-sample regression target would pull predictions toward the conditional mean. Specifically, a frozen V-JEPA~2 encoder~\cite{assran2025vjepa2} independently encodes each camera view. We concatenate the features along the channel dimension while preserving the full spatiotemporal token grid. We denote this feature extraction process by $F$ and channel-wise standardization by $S$, using fixed means and standard deviations estimated from reference data. The resulting representations are $z=S(F(\tau))$ and $\hat z=S(F(\hat\tau))$.

A conditional critic $D_\psi$ compares these representations under the shared condition $c=(h(o_0),g,a_{0:K-1})$, where $\psi$ denotes its parameters. The initial observation feature $h(o_0)$ is obtained by repeating each initial-view image over time, encoding it, averaging over all tokens, and concatenating the per-view vectors. The critic assigns a conditional trajectory score and minimizes a Wasserstein-style \mbox{objective~\cite{arjovsky2017wgan,gulrajani2017wgangp,chemseddine2025conditional}:}
\begin{equation}
\mathcal L_D(\psi)
= \tfrac{1}{s}
\bigl(\mathbb E[D_\psi(\hat z,c)]
- \mathbb E[D_\psi(z,c)]\bigr)
+ \lambda_{\mathrm{gp}}\mathcal L_{\mathrm{gp}}.
\label{eq:critic}
\end{equation}
Here, the expectations cover reference trajectories and their generated counterparts; $s=\sqrt d$ is a scale factor, with $d$ the total number of elements in the concatenated trajectory feature. The term $\mathcal L_{\mathrm{gp}}$ is a gradient penalty evaluated at interpolations between reference and generated representations, and $\lambda_{\mathrm{gp}}$ is its weight. This penalty approximately enforces the critic's Lipschitz constraint; its definition is \mbox{provided in Appendix~B.}

\paragraph{Training objective.}
To retain both trajectory-level supervision and local generative constraints, we use conditional trajectory alignment as the primary objective and self-rollout generative supervision as an auxiliary term:
\begin{equation}
\mathcal L_{\mathrm{world}}(\phi)
=
-\frac{\lambda_{\mathrm W}}{s}
\mathbb E[D_\psi(\hat z_\phi,c)]
+
\mathcal L_{\mathrm{gen}}^{\mathrm{SR}}(\phi).
\label{eq:pda}
\end{equation}
Here, $\hat z_\phi=S(F(\hat\tau_\phi))$, with the subscript emphasizing the dependence of the generated trajectory and its representation on $\phi$. The coefficient $\lambda_{\mathrm W}$ controls the alignment weight relative to auxiliary supervision. By increasing the critic's score for generated trajectories, the world model learns to match the reference representation distribution under the same conditions. During world-model updates, the visual VAE, video encoder, and critic parameters remain frozen, while alignment gradients propagate through trajectory decoding and feature extraction to the world model. Observations passed between chunks are detached to \mbox{control training cost.}

\begin{table*}[!b]
    \centering
    \caption{\textbf{World-model prediction quality, action-sequence ranking, and object-center localization.}
    Training chunks specifies the autoregressive post-training horizon (--: not applicable); all methods predict 240 frames at evaluation. AUROC uses complete trajectories. Object-center error measures Euclidean distance in $256\times256$ front-view images, averaged first over frames and then over trajectories. GroundedDream uses four training chunks by default. Bold: best; underline: second best.}
    \label{tab:world_model_quality}

    \begingroup
    \footnotesize
    \setlength{\tabcolsep}{1.5pt}
    \renewcommand{\arraystretch}{1.0}
    \begin{tabular*}{\textwidth}{@{\extracolsep{\fill}}lc*{12}{c}@{}}
        \toprule
        \multirow{2}{*}{Method}
        & \multirow{2}{*}{\shortstack{Training\\chunks}}
        & \multicolumn{5}{c}{Front View}
        & \multicolumn{5}{c}{Wrist View}
        & \multirow{2}{*}{AUROC$\uparrow$}
        & \multirow{2}{*}{\shortstack{Object-Center\\Error (px)$\downarrow$}} \\
        \cmidrule(lr){3-7}
        \cmidrule(lr){8-12}
        & &
        LPIPS$\downarrow$ & FID$\downarrow$ & FVD$\downarrow$
        & SSIM$\uparrow$ & PSNR$\uparrow$
        & LPIPS$\downarrow$ & FID$\downarrow$ & FVD$\downarrow$
        & SSIM$\uparrow$ & PSNR$\uparrow$ & & \\
        \midrule
        Unaligned world model & --
        & 0.103 & 20.17 & 77.78 & 0.826 & 22.07
        & 0.339 & 70.98 & 245.65 & 0.659 & 17.30 & 0.712 & 9.02 \\
        Ctrl-World & --
        & 0.134 & 26.50 & 82.40 & 0.803 & 20.44
        & 0.335 & 54.09 & 301.21 & 0.684 & 16.85 & 0.610 & 9.03 \\
        \midrule
        \multirow{3}{*}{Self Forcing (DMD)} & 2
        & 0.103 & 11.52 & 105.63 & 0.806 & 21.26
        & 0.369 & 40.02 & 160.56 & 0.615 & 17.23 & 0.685 & 8.65 \\
        & 4
        & 0.094 & 10.51 & 98.13 & 0.826 & 22.30
        & 0.326 & 37.83 & 143.57 & 0.635 & 17.52 & 0.699 & 8.23 \\
        & 8
        & 0.098 & 10.98 & 58.94 & 0.815 & 22.23
        & 0.345 & 38.56 & 144.26 & 0.642 & 17.48 & 0.702 & 8.17 \\
        \midrule
        \multirow{3}{*}{GroundedDream (ours)} & 2
        & 0.082 & 9.02 & \underline{42.12} & 0.829 & 23.12
        & \underline{0.203} & \underline{32.36}
        & \underline{92.36} & \underline{0.703}
        & \underline{18.29} & 0.823 & 6.15 \\
        & 4
        & \textbf{0.061} & \underline{8.96} & \textbf{35.77}
        & \textbf{0.862} & \textbf{23.94}
        & \textbf{0.198} & \textbf{30.80} & \textbf{85.64}
        & \textbf{0.710} & \textbf{18.69} & \textbf{0.849} & \textbf{5.47} \\
        & 8
        & \underline{0.069} & \textbf{8.11} & 57.21
        & \underline{0.851} & \underline{23.44}
        & 0.213 & 54.21 & 191.67 & 0.695 & 18.24
        & \underline{0.837} & \underline{6.03} \\
        \bottomrule
    \end{tabular*}
    \endgroup
\end{table*}

\subsection{Policy Evaluation and Optimization with GroundedDream}
\label{sec:policy_imagination}

\paragraph{Chunk-wise closed-loop rollout.}
After PDA, we freeze the world model $M_{\phi^*}$ as an imagined environment for policy $\pi_\theta$, parameterized by $\theta$ (Fig.~\ref{fig:method}b). Starting from $\hat o_0=o_0$ and instruction $g$, the policy generates an action chunk $a_k\sim\pi_\theta(\cdot\mid\hat o_k,g)$, and the world model predicts the subsequent observation chunk $\hat O_{k+1}\sim M_{\phi^*}(\cdot\mid\hat o_k,g,a_k)$. Its last frame $\hat o_{k+1}$ becomes the next policy observation. Repeating this process for $J$ action chunks produces $\hat\tau=(\hat O_1,\ldots,\hat O_J)$, where $J$ is the policy rollout horizon.

\paragraph{Deployment-free policy evaluation.}
An independently trained, frozen reward model scores temporal windows from each complete rollout. We average the highest-scoring windows to obtain a trajectory score $q(\hat\tau,g)$. The reward model conditions on both the rollout video and the task instruction. We define the terminal reward and predicted success rate as
\begin{equation}
\begin{aligned}
R(\hat\tau,g)
&=r_+\,\mathbf 1[q(\hat\tau,g)\geq\kappa],\\
\hat S(\pi_\theta)
&=\frac{1}{N}\sum_{n=1}^{N}
\mathbf 1[q(\hat\tau^{(n)},g^{(n)})\geq\kappa].
\end{aligned}
\label{eq:eval}
\end{equation}
Here, $\kappa$ is a fixed success threshold, $r_+>0$ is the success reward, and $\mathbf 1[\cdot]$ is the indicator function. The number of evaluation conditions is $N$, and $\hat\tau^{(n)}$ is the rollout from the $n$-th initial observation and instruction $g^{(n)}$. All policies are evaluated under the same conditions for performance comparison and checkpoint selection.

\paragraph{Policy optimization with trajectory-level feedback.}
For each initial condition, we sample $G$ trajectories and compute group-relative advantages from terminal rewards as $A_i=(R_i-\bar R)/(\sigma_R+\eta)$. Here, $i$ indexes trajectories within the group, $R_i=R(\hat\tau_i,g)$ is the terminal reward, $\bar R$ and $\sigma_R$ are the group reward mean and standard deviation, and $\eta>0$ ensures numerical stability. Following Group Relative Policy Optimization (GRPO)~\cite{shao2024deepseekmath} and Flow-GRPO~\cite{liu2025flowgrpo}, we optimize stochastic action-denoising transitions using
\begin{equation}
\begin{gathered}
\mathcal L_{\mathrm{policy}}(\theta)
= -\mathbb E_{i,k,j}\Bigl[
\min\bigl(\rho_{i,k,j}A_i,\\
\operatorname{clip}(\rho_{i,k,j},1-\delta,1+\delta)A_i
\bigr)\Bigr].
\end{gathered}
\label{eq:policy}
\end{equation}
Here, $k$ and $j$ index action chunks and denoising transitions, respectively. The ratio
$\rho_{i,k,j}=\exp[\ell_\theta(i,k,j)-\ell_{\theta_{\mathrm{old}}}(i,k,j)]$
compares the probability of the same sampled transition under the current and behavior policies. The quantity $\ell_\theta$ is its log-probability given the corresponding observation, instruction, and denoising conditions, and $\theta_{\mathrm{old}}$ denotes the policy parameters used to collect the rollouts. Here, $\delta>0$ is the clipping threshold. The expectation averages over sampled trajectories and their action-denoising transitions. All transitions within a trajectory share its terminal advantage, avoiding a learned value network.

\paragraph{Joint optimization of visual and action components in WAMs.}
For WAMs, action-transition log-probabilities are recomputed from cached video and action latents, which are treated as constants. Gradients propagate through the visual input projection, shared generative backbone, and action-generation modules participating in this recomputation. The visual VAE and imagined environment remain frozen, and gradients do not propagate through earlier cached denoising transitions. Thus, terminal advantages optimize action-transition likelihoods through the policy's visual and action components, without differentiating through the environment or introducing a separate \mbox{video-prediction objective.}

\section{Experiments}
\label{sec:experiments}

\begin{figure*}[!tb]
    \centering
    \includegraphics[width=\textwidth]{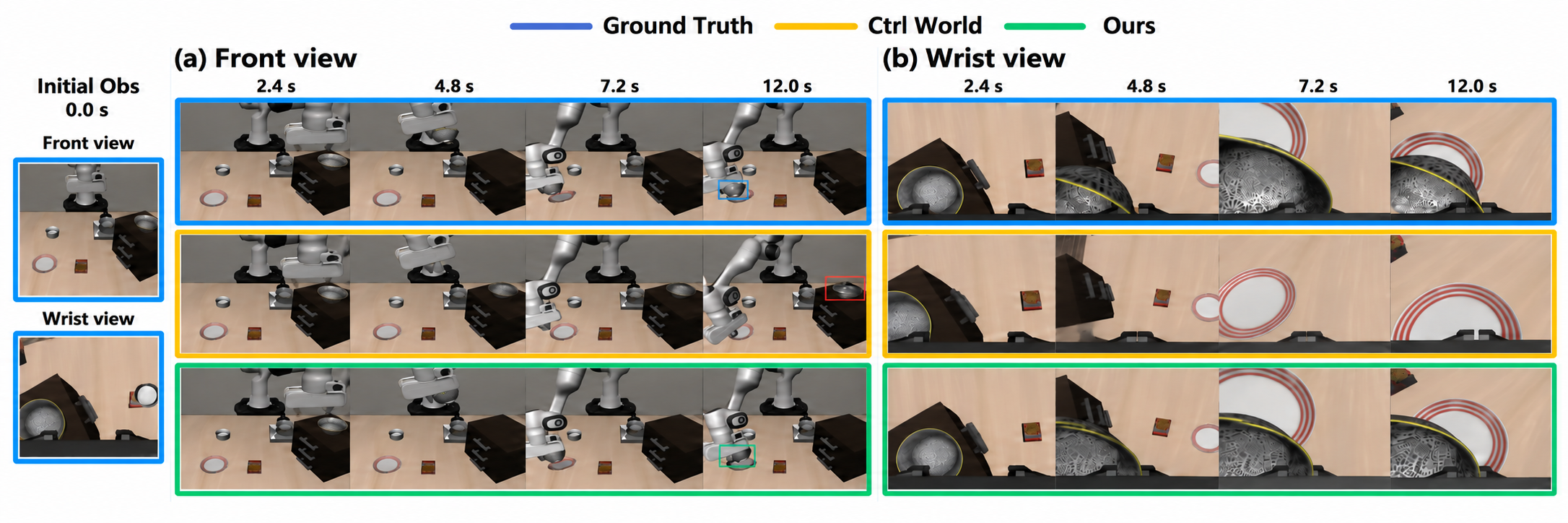}
    \caption{\textbf{Long-horizon action-conditioned prediction.}
    Under the same initial observation and action sequence, we compare Ctrl-World and GroundedDream with simulator ground truth at 2.4, 4.8, 7.2, and 12.0\,s. The left and right panels show the front and wrist views, respectively. GroundedDream more closely preserves the arm position and the wrist-camera view at later time steps.}
    \label{fig:action_outcomes}
\end{figure*}

\begin{table*}[t]
  \centering
  \caption{\textbf{Ablation of PDA.} ``w/o Trajectory Alignment'' removes conditional Wasserstein alignment but retains self-rollout flow matching. ``w/ Chunk-wise Alignment'' averages independent chunk-level alignment losses over the same four-chunk rollout instead of aligning it jointly. ``w/o Critic Action Cond.'' removes action conditioning from the critic only. The last two variants replace frozen V-JEPA~2 features with VideoMAE features or VAE latents, retaining trajectory alignment. Metrics follow Table~\ref{tab:world_model_quality}. Bold: best; underline: second best.}
  \label{tab:ablation_cvpr}
  \begingroup
  \footnotesize
  \setlength{\tabcolsep}{1.5pt}
  \renewcommand{\arraystretch}{1.15}
  \begin{tabular*}{\textwidth}{@{\extracolsep{\fill}}l*{11}{r}c@{}}
    \toprule
    \textbf{Method}
    & \multicolumn{5}{c}{\textbf{Front View}}
    & \multicolumn{5}{c}{\textbf{Wrist View}}
    & \multirow{2}{*}{\textbf{AUROC}$\uparrow$}
    & \multirow{2}{*}{\shortstack{\textbf{Object-Center}\\\textbf{Error (px)}$\downarrow$}} \\
    \cmidrule(lr){2-6}\cmidrule(lr){7-11}
    & LPIPS$\downarrow$ & FID$\downarrow$ & FVD$\downarrow$ & SSIM$\uparrow$ & PSNR$\uparrow$
    & LPIPS$\downarrow$ & FID$\downarrow$ & FVD$\downarrow$ & SSIM$\uparrow$ & PSNR$\uparrow$
    & & \\
    \midrule
    \textbf{GroundedDream (Full)}
    & \textbf{0.061} & \textbf{8.96} & \textbf{35.77} & \textbf{0.862} & \textbf{23.94}
    & \textbf{0.198} & \textbf{30.80} & \textbf{85.64} & \textbf{0.710} & \textbf{18.69}
    & \textbf{0.849} & \textbf{5.47} \\
    \midrule
    w/o Trajectory Alignment
    & 0.072 & 19.50 & 42.59 & 0.840 & 22.79
    & 0.205 & 33.37 & 98.88 & \underline{0.707} & 17.58
    & 0.791 & 8.06 \\
    w/ Chunk-wise Alignment
    & \underline{0.063} & 12.96 & 40.36 & 0.848 & 22.84
    & \underline{0.203} & 33.28 & \underline{92.34} & 0.704 & 17.63
    & 0.816 & 7.95 \\
    w/o Critic Action Cond.
    & 0.066 & \underline{12.87} & 39.65 & \underline{0.852} & \underline{23.01}
    & 0.207 & \underline{31.22} & 100.96 & 0.671 & 17.61
    & \underline{0.835} & \underline{6.98} \\
    w/o V-JEPA 2 (w/ VideoMAE)
    & 0.077 & 13.93 & \underline{39.53} & 0.839 & 22.97
    & 0.221 & 42.78 & 197.07 & 0.687 & \underline{17.98}
    & 0.834 & 7.56 \\
    w/o V-JEPA 2 (w/ VAE latent)
    & 0.107 & 27.93 & 195.47 & 0.818 & 22.99
    & 0.304 & 73.90 & 365.79 & 0.686 & 17.91
    & 0.793 & 7.43 \\
    \bottomrule
  \end{tabular*}
  \endgroup
\end{table*}

\begin{figure}[!tbp]
    \centering
    \includegraphics[width=\linewidth]{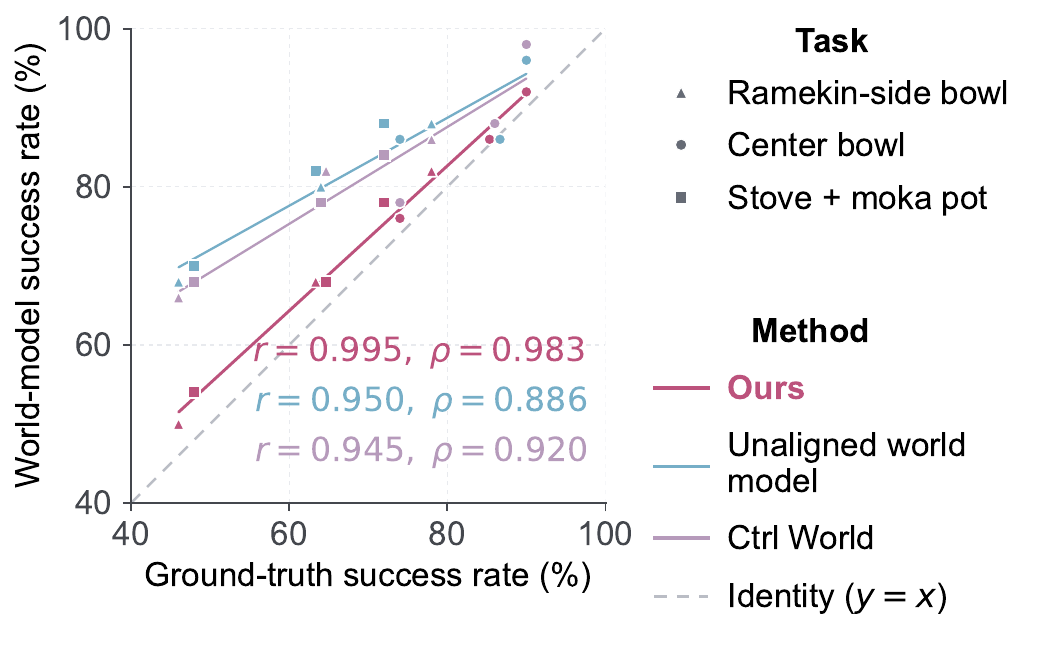}
    \caption{\textbf{Consistency between imagined and simulator policy evaluation (WAM).}
    Each point is one task--checkpoint pair (markers identify tasks), comparing the success rate estimated in a world model with that measured in the simulator. The dashed line denotes ideal agreement ($y=x$) and solid lines are linear fits; Pearson correlation $r$ and Spearman rank correlation $\rho$ are reported across the plotted task--checkpoint pairs. Rollouts last 12\,s for the placement tasks and 26\,s for the stove-and-moka-pot task.}
    \label{fig:success_consistency}
\end{figure}

We evaluate GroundedDream through three \mbox{research questions:}
\begin{itemize}
    \item \textbf{(Q1) Visual quality and physical fidelity.}
    To what extent does PDA improve long-horizon prediction quality and bring predicted arm motion and scene changes closer to simulator trajectories?

    \item \textbf{(Q2) Policy evaluation.}
    Does PDA make imagined rollouts a more accurate evaluator of VLA and WAM policies?

    \item \textbf{(Q3) Policy optimization.}
    Does PDA make reinforcement learning inside the world model more effective at improving the success rates of VLA and WAM policies in the simulator and on a real robot?
\end{itemize}

\subsection{(Q1): Visual quality and physical fidelity of long-horizon prediction}
\label{sec:world_model_quality}

\paragraph{Prediction quality and action-sequence ranking.}
Visual fidelity alone does not guarantee that a world model is useful for policy learning: its predictions become inputs to subsequent decisions and must therefore preserve information that distinguishes successful from unsuccessful actions. We evaluate video prediction quality in both front and wrist views, together with action-sequence ranking. The latter is measured by AUROC, computed from reward scores on generated trajectories against simulator success labels. We additionally measure object-center localization error to assess spatial agreement between generated and \mbox{reference trajectories.}

We compare GroundedDream with the Unaligned world model, Ctrl-World, and Self Forcing with distribution matching distillation (DMD) on LIBERO-Spatial. All post-training methods use the same training data, and all methods are evaluated using identical initial observations and recorded action sequences. The Unaligned world model is the initial world model before PDA post-training. We also conduct extensive evaluations on other datasets; these results, together with data, training, and evaluation details, are provided in Appendices~C and~\ref{app:long-prediction}.

Table~\ref{tab:world_model_quality} shows that GroundedDream achieves the best results on nearly all metrics, notably improving FVD and AUROC while reducing object-center localization error from 9.02 to 5.47\,px, a 39.4\% reduction. Although Self Forcing (DMD) improves FID, its FVD varies across views and training chunk counts, and its AUROC remains below that of the Unaligned world model. GroundedDream instead improves video prediction, action-sequence ranking, and object localization together, supporting the effectiveness of conditional trajectory alignment. With the evaluation horizon fixed, extending the training rollout from $K=4$ to $K=8$ brings no further gain. AUROC and object-center error remain comparable (0.837 vs.\ 0.849 and 6.03 vs.\ 5.47\,px) and front-view FID improves slightly, whereas FVD rises in both views. Four chunks thus appear sufficient to capture the physical dynamics of these tasks, and we use $K=4$ in subsequent experiments.

\begin{table*}[!t]
    \makeatletter\def\@captype{figure}\makeatother
    \centering
    \includegraphics[width=\textwidth]{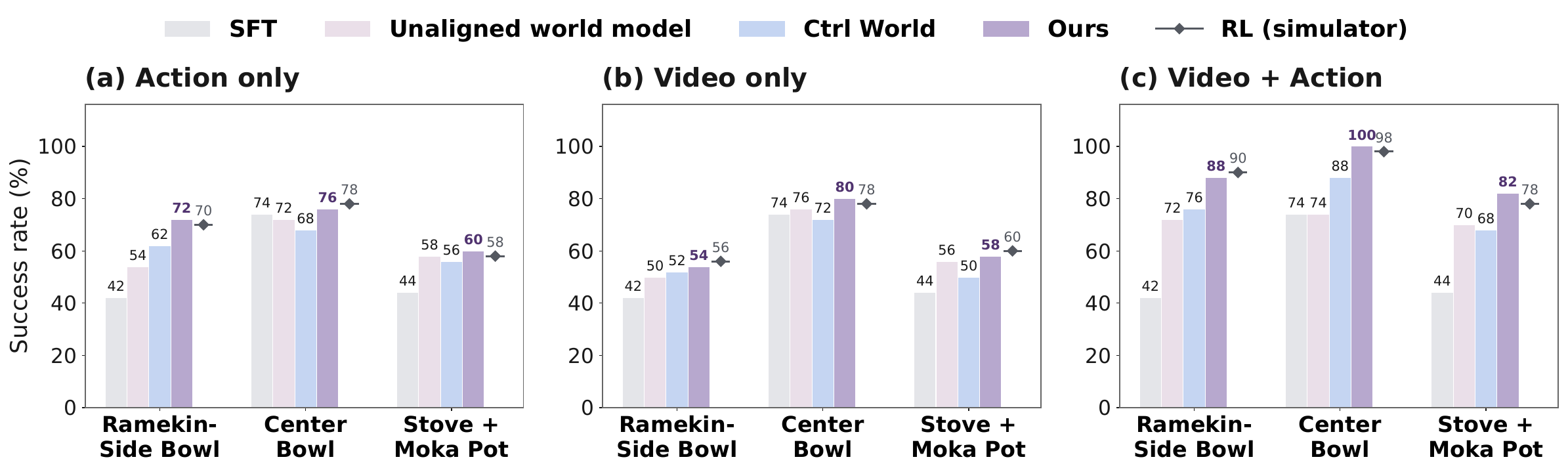}
    \caption{\textbf{Policy optimization on LIBERO (WAM).}
    Simulator success rates (\%) after RL in each world model under three update scopes: (a) action-interface layers only, (b) video-generation layers only, and (c) both jointly. SFT is the shared initial policy. Each task is evaluated on 50 placements. Gray diamonds indicate policy success rates after RL directly in the simulator.}
    \label{fig:policy_improvement}
\end{table*}

\paragraph{Long-horizon action-outcome prediction.}
To examine when and where predictions deviate from actual action consequences, we compare generated observations with simulator trajectories under identical initial observations and action sequences (Fig.~\ref{fig:action_outcomes}). The front view shows arm and object motion, while the wrist view shows local changes around the gripper. Ctrl-World gradually drifts from the reference arm configuration, whereas GroundedDream maintains agreement with the reference in arm position and wrist-camera view until the end of the rollout.

\paragraph{Ablation of trajectory alignment, action conditioning, and latent space.}
Table~\ref{tab:ablation_cvpr} isolates three aspects of PDA: joint trajectory alignment, action conditioning, and the predictive latent space. Policy rollout data and self-rollout training alone already improve action-sequence ranking. Adding trajectory alignment further improves AUROC and accounts for most of the gains in object localization accuracy and front-view visual quality. Replacing joint alignment with independent chunk-wise alignment also reduces prediction quality, action-sequence ranking, and object localization accuracy. These results support supervising action consequences jointly across multiple \mbox{prediction chunks.}

Removing action conditioning from the critic weakens action-sequence ranking and object localization, while replacing V-JEPA~2 with VideoMAE~\cite{tong2022videomae} features or VAE latents also degrades performance. Together, these comparisons support aligning generated and reference trajectories jointly in a predictive latent space under matched \mbox{action conditions.}

\begin{figure}[!t]
    \centering
    \includegraphics[width=\linewidth]{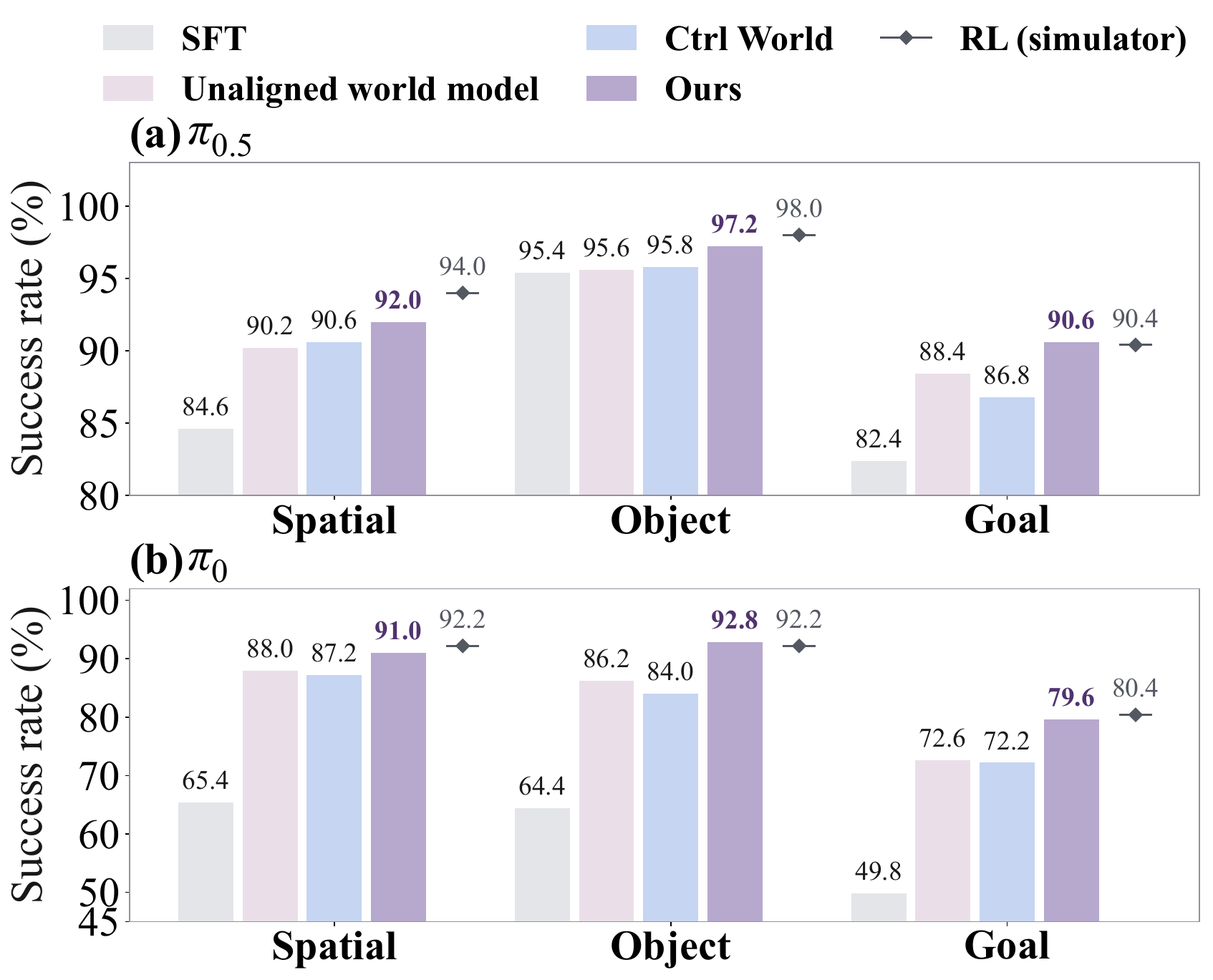}
    \caption{\textbf{VLA policy optimization on LIBERO.}
    Simulator success rates (\%) for (a) $\pi_{0.5}$ and (b) $\pi_0$ on Spatial, Object, and Goal. SFT is the initial policy for each backbone; the remaining bars apply RL in the Unaligned world model, Ctrl-World, or GroundedDream (Ours). Gray diamonds indicate policy success rates after RL directly in the simulator. Panels use different \mbox{vertical-axis ranges.}}
    \label{fig:vla_policy_improvement}
\end{figure}

\subsection{(Q2): Agreement between imagined and simulator policy evaluation}
\label{sec:policy_evaluation}

Low per-step error does not by itself guarantee that a world model ranks policies correctly, since evaluation depends on whether complete closed-loop executions succeed or fail. Moreover, a world model trained mainly on successful demonstrations can remain overly optimistic when a policy deviates from them, masking differences between weaker and stronger policies. We therefore roll out multiple checkpoints of the Unified WAM policy built on Cosmos~3 Edge~\cite{agarwal2026cosmos3} in closed loop inside each world model and in the simulator from matched initial conditions, and compare the resulting success rates. The two spatial placement tasks assess differences between policies in object placement, while the sequential stove-and-moka-pot task tests whether agreement holds over a longer sequence of operations.

As shown in Fig.~\ref{fig:success_consistency}, the Unaligned world model and Ctrl-World systematically overestimate success, especially for weaker checkpoints. This flattens their fitted lines and compresses the gaps between policies. In contrast, GroundedDream estimates are closer to the identity line across all three tasks and achieve the highest Pearson and Spearman correlations across the evaluated task--checkpoint pairs. These results show closer agreement with simulator evaluation and support using imagined rollouts to compare policies. We conduct the same evaluation for the VLA policies $\pi_{0.5}$~\cite{pi2025pi05} and $\pi_0$~\cite{black2024pi0}, with results \mbox{provided in Appendix~E.2.}

\begin{figure}[!t]
    \centering
    \includegraphics[width=\linewidth]{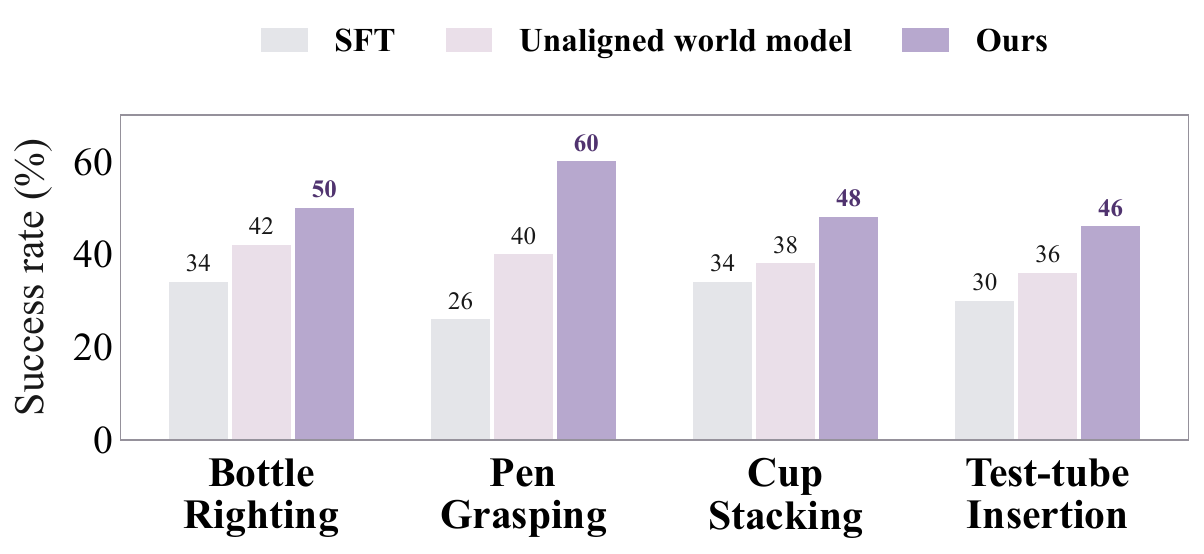}
    \caption{\textbf{Real-robot success rates (\%) (WAM).}
    SFT is the initial policy; ``Unaligned world model'' and ``Ours'' apply RL in the world model before and after PDA, respectively. Each task is evaluated over 50 trials.}
    \label{fig:real_robot}
\end{figure}

\subsection{(Q3): Policy optimization through reinforcement learning in the world model}
\label{sec:policy_improvement}

We test whether improved world-model fidelity translates into policy gains. Each initial policy is obtained by SFT on one demonstration per task, which leaves headroom for RL. Within each policy backbone, all world-model comparisons use the same initial SFT policy and reward model. The total number of training rounds and the number of optimizer steps are identical across the compared world models.

\paragraph{Policy optimization for WAMs and VLAs.}
We optimize the Cosmos~3 Edge Unified WAM and the VLA policies $\pi_{0.5}$ and $\pi_0$ with RL inside each world model. GroundedDream achieves the highest success rates among the compared world models across all evaluated tasks and WAM update scopes (Fig.~\ref{fig:policy_improvement}), as well as for both VLA backbones across all three LIBERO suites (Fig.~\ref{fig:vla_policy_improvement}). For the WAM, jointly updating the visual and action components consistently outperforms updating either alone. In this setting, GroundedDream raises the average success rate from 53.3\% after SFT to 90.0\%. Across these settings, policies optimized in GroundedDream achieve success rates close to those obtained through simulator-based RL, supporting its effectiveness as an imagined environment for \mbox{policy optimization.}

\paragraph{Policy optimization on a real robot.}
We evaluate the Cosmos~3 Edge Unified WAM on the AgileX dual-arm platform. The policy is initialized by SFT on a small set of demonstrations, and the world model is trained on 80 real-robot interaction trajectories. As shown in Fig.~\ref{fig:real_robot}, GroundedDream improves success rates on bottle righting, pen grasping, cup stacking, and test-tube insertion relative to both the initial SFT policy and policies optimized in the Unaligned world model, with the largest gain on pen grasping. Averaged across the four tasks, GroundedDream improves real-robot success by 20.0 percentage points over SFT and by 12.0 points over RL in the Unaligned world model. Detailed optimization results and real-robot analyses are provided in Appendices~E.3 and~E.5, respectively.

\section{Conclusion}
\label{sec:conclusion}

We presented GroundedDream, a post-training framework that improves the physical fidelity of action-conditioned world models by aligning generated and reference trajectory distributions in a predictive latent space. Experiments demonstrate that GroundedDream improves long-horizon prediction, makes policy evaluation less overoptimistic, and supports effective policy optimization in imagination. Across four real-robot tasks, policies optimized in GroundedDream achieve an average absolute success-rate gain of 20.0\% over the initial SFT policy.

Despite these promising results, several important challenges remain. GroundedDream relies on reference interaction data for trajectory alignment, and its generalization to tasks and interaction states beyond the coverage of these data requires further evaluation. Moreover, the world model remains fixed during policy optimization; as the policy evolves, it may visit states beyond the world model's post-training distribution. Future work could explore active collection of interaction data and alternating updates of the world model and policy to expand coverage of task-relevant states and support continued policy improvement.

\clearpage
\section{LLM Usage Statement}
We used GPT-series LLMs for manuscript language editing and code implementation and debugging. All use was directed and supervised by the authors, who reviewed all edits and rigorously checked the assisted code. The models did not conduct autonomous research. 
The authors take full responsibility for the paper and code.

\clearpage
\appendix
\maketitlesupplementary

This supplement provides implementation details, experimental settings, and evaluation protocols for the GroundedDream method and results presented in the main paper. Appendix~A expands on the related work, and Appendix~B details trajectory alignment and policy optimization in imagination. Appendix~C describes reward-model training, threshold calibration, and trajectory scoring, and reports classification accuracy on held-out simulator videos. The supplementary prediction experiments referenced in the main paper are in Appendix~D. Appendix~\ref{app:gd-world-training} gives the world-model training and evaluation settings and the paired regression control, Appendix~\ref{sec:physical_event_evaluation} examines grasp and transport events, and Appendix~\ref{app:long-prediction} reports the LIBERO-Long results. Appendix~E details the closed-loop policy evaluation and optimization protocols and reports results under initial placement shifts, across independent training seeds, and for each real-robot task.

\section{Extended Related Work}
\label{app:related_work}

\paragraph{Action-conditioned video models for interaction.}
Action-conditioned world models predict future observations from the current scene and the actions applied to it~\cite{wu2024ivideogpt,guo2025ctrlworld}. In Ctrl-World~\cite{guo2025ctrlworld}, generated camera observations can become the inputs to a visual policy at its next decision. An error that occupies only a few pixels can therefore change the course of an interaction. If the model depicts an object rising with the gripper after a missed grasp, the policy may initiate transport because its observation suggests that the grasp succeeded. This setting gives a concrete reference for assessing a prediction: what happened when those actions were actually executed? GroundedDream retains the video interface used by the policy and pairs each training rollout with a recorded execution under the same initial observation, instruction, and action sequence.

\paragraph{Predictive representations as supervision for dynamics.}
Pretrained video features offer ways to compare motion and interaction beyond frame reconstruction. VideoREPA~\cite{zhang2025videorepa}, for example, transfers relational structure from a foundation model to a video generator. LDO~\cite{luo2026ldo} uses a frozen predictive model to align the structure of generator features and derives a reinforcement learning reward from temporal predictability. In robotic manipulation, MIND-V~\cite{zhang2025mindv} uses V-JEPA~2~\cite{assran2025vjepa2} to construct a foresight-coherence reward for video generation. GroundedDream also uses a frozen predictive encoder, but the training reference is a recorded consequence of the particular actions supplied to the generator. For a grasp followed by a lift, the recording shows whether the object moved with the gripper and whether it stayed in the grasp during transport. Encoding both executions in the same space lets a conditional critic use discrepancies anywhere along that sequence.

\paragraph{Autoregressive rollout and post-training objectives.}
Self Forcing~\cite{huang2025selfforcing} conditions the generator on its own previous outputs during training, exposing it to the histories it will encounter at inference. Adversarial post-training (APT)~\cite{lin2025apt} uses a discriminator to improve one-step video generation. In world-model post-training, RLVR-World~\cite{wu2025rlvrworld} uses rewards based on prediction quality, while RLIR~\cite{ye2025rlir} uses an inverse model to assess whether the actions underlying a generated transition can be recovered. For manipulation, the most informative evidence about one transition may appear several transitions later. At gripper closure, the fingers may occlude the contact region; once the arm lifts, the subsequent motion reveals whether the object was secured. PDA unrolls several chunks along the recorded actions and assigns a joint score to their predictive features. The score can thus depend on both closure and transport, while the flow-matching objective continues to supervise the visual predictions within each chunk.

\paragraph{Policy evaluation and optimization in imagination.}
WorldGym~\cite{quevedo2025worldgym} evaluates policies using action-conditioned video rollouts. Ctrl-World~\cite{guo2025ctrlworld} also evaluates policies through imagined interaction and uses selected successful synthetic trajectories for supervised policy improvement. WMPO~\cite{zhu2025wmpo} adapts a world model with policy execution data before optimizing a VLA through imagined rollouts. WoVR~\cite{jiang2026wovr} studies the reliability of such updates when the world model is imperfect. Policy evaluation and optimization place distinct demands on an imagined environment. For evaluation, an estimated success rate should track differences between policy checkpoints. During optimization, an action sequence rewarded in imagination must still succeed when the updated policy is executed. GroundedDream freezes the aligned world model before policy learning. We examine these two demands separately by comparing imagined and simulator success rates for matched policies, then executing the optimized policies in the simulator and on the robot.

\section{Additional Method Details}
\label{app:method_details}
GroundedDream aligns conditional trajectory distributions to improve action-outcome prediction while retaining image generation for policy evaluation and optimization. Training combines two complementary paths: self-rollout flow matching supervises reference futures under generated histories, while a conditional critic jointly assesses consecutive observation chunks in a predictive representation space. After alignment, the world model is frozen as an imagined environment, and policies are optimized using terminal feedback from complete rollouts.

This appendix supplements Secs.~3.1--3.2 and Eqs.~(1)--(5) of the main paper. Reward-model data, world-model training settings, and policy optimization budgets are detailed in Appendices~\ref{app:reward-blind}, \ref{app:gd-world-training}, and~\ref{app:rl-settings-results}, respectively. Following the main paper, $\phi$, $\psi$, and $\theta$ parameterize the world model, critic, and policy, respectively; $K$ is the world-model training horizon in chunks, whereas $J$ is the closed-loop policy horizon. Hats denote generated observations, trajectories, or representations.

\subsection{World-Model Interface and Action Conditioning}
\label{app:world_interface}
Closed-loop interaction requires predicting future observations from the current observation and intended actions. We use a chunk-based interface in which action and observation chunks cover the same physical interval, and generated observations are passed between calls.

\paragraph{Conditional generation.}
We initialize the world model from Cosmos 3 Edge, retaining its visual VAE and multimodal generative backbone~\cite{cosmosedge}. Let $\hat o_k$ be the current observation, $g$ the instruction, and $a_k$ an ordered action chunk. A call produces
\begin{equation}
\hat O_{k+1}=M_\phi(\operatorname{sg}(\hat o_k),g,a_k;\xi_k),
\qquad \hat o_0=o_0.
\end{equation}
Here $o_0$ is the real initial observation, $\xi_k$ denotes sampling randomness, and $\operatorname{sg}$ is stop-gradient. The last generated frame, after observation preprocessing, forms $\hat o_{k+1}$. The conditioning frame is not counted again among newly predicted frames.

\paragraph{Views and actions.}
Front and wrist images are concatenated horizontally in a fixed order and jointly generated in one call. Actions preserve their within-chunk order and enter the multimodal sequence through a learned projection and modality embedding. Known action conditions are distinguished from noisy policy actions, which additionally receive denoising-time embeddings. Policy outputs are converted to executed actions and then to the world-model conditioning representation. Appendix~\ref{app:gd-world-training} specifies the control frequency and the number of actions per chunk.

\subsection{Multi-Chunk Self-Rollout and Reference Supervision}
\label{app:self_rollout}

\paragraph{Consecutive rollout.}
We sample $o_0$, consecutive actions $a_{0:K-1}$, and corresponding real observation chunks $O_{1:K}$. The generated terminal observation conditions the next chunk:
\begin{equation}
\hat o_{k+1}=\operatorname{Boundary}(\hat O_{k+1}).
\end{equation}
The boundary operator includes last-frame extraction, range conversion, and observation preprocessing. Its output is detached in the next call, while the current generated chunk retains its trajectory-alignment path.

\paragraph{Reference latents.}
The frozen VAE supplies $Z_{k+1}$ in main-paper Eq.~(1). Encoding includes the current boundary and the corresponding real future:
\begin{equation}
Z_{k+1}=E_{\mathrm{VAE}}
\left([\operatorname{sg}(\hat o_k);O_{k+1}]\right).
\end{equation}
Here $E_{\mathrm{VAE}}$ is frozen and the semicolon denotes temporal concatenation. The first chunk uses a real boundary and subsequent chunks use generated boundaries; future observations always come from the reference trajectory.

\paragraph{Condition-preserving flow matching.}
We regress the velocity of a linear data--noise path~\cite{lipman2023flow}. Let $\sigma$ be the noise fraction and $\epsilon\sim\mathcal N(0,I)$ be shape-matched Gaussian noise. With $M_k$ equal to one at fixed conditioning positions and zero elsewhere, the packed input is
\begin{equation}
\begin{aligned}
x_{k,\sigma}^{\mathrm{pack}}
=&M_k\odot Z_{k,\mathrm{cond}}\\
&+(1-M_k)\odot[(1-\sigma)Z_{k+1}+\sigma\epsilon].
\end{aligned}
\end{equation}
Here $\odot$ denotes elementwise multiplication and $Z_{k,\mathrm{cond}}$ contains layout-compatible conditioning values. Future positions follow $x_\sigma$ in the main paper, while conditioning positions remain fixed. The same $Z_{k+1}$ is used for both noising and velocity supervision.
\begin{equation}
\begin{aligned}
\mathcal L_{\mathrm{gen}}^{\mathrm{SR}}
=\mathbb E\Bigg[\frac1K\sum_{k=0}^{K-1}
\operatorname{mean}_{\mathrm{future}}\!\Big[ \Big(
&v_\phi(x_{k,\sigma}^{\mathrm{pack}},\sigma\mid
\operatorname{sg}(\hat o_k),g,a_k)\\
&-(\epsilon-Z_{k+1})\Big)^2\Big]\Bigg].
\end{aligned}
\end{equation}
The mean covers valid future visual-latent scalars, excluding conditioning positions, actions, and text. This specifies the implemented reduction of the squared error in main-paper Eq.~(1).

\paragraph{Sampling gradients.}
Generation uses numerical integration. To control cost, we retain gradients only through the final integration update of each chunk; earlier sampling updates and inter-chunk boundaries are detached. Every chunk can therefore receive trajectory supervision without differentiating through the complete sampling chain or recursive history. Appendix~\ref{app:gd-world-training} gives the Euler integration grid.

\subsection{Predictive Trajectory Representation and Conditional Critic}
\label{app:critic}
Trajectory alignment requires retaining temporal changes and assessing relationships across chunk boundaries. We extract spatiotemporal features and process them with a critic that connects time steps and views.

\paragraph{Features and standardization.}
A frozen V-JEPA 2 encoder~\cite{assran2025vjepa2} independently encodes the complete training window of each view. Final-layer spatiotemporal tokens are concatenated across channels. Temporal indices identify synchronized observations; spatial indices do not imply geometric correspondence across cameras. Following the main paper,
\begin{equation}
z=S(F(\tau)),\qquad \hat z=S(F(\hat\tau)),
\end{equation}
where $\tau=(O_1,\ldots,O_K)$, $\hat\tau=(\hat O_1,\ldots,\hat O_K)$, $F$ extracts and concatenates features, and $S$ applies fixed channel-wise standardization. For channel $r$ and spatiotemporal position $p$,
\begin{equation}
S(F)_{r,p}=\frac{F_{r,p}-\mu_r}
{\sqrt{\max(\nu_r,\epsilon_{\mathrm{var}})}}.
\end{equation}
The means $\mu_r$ and variances $\nu_r$ are estimated over training reference samples and their temporal and spatial positions; $\epsilon_{\mathrm{var}}>0$ is a variance floor. Real and generated features use the same fixed statistics.

\paragraph{Conditions and joint scoring.}
We retain $c=(h(o_0),g,a_{0:K-1})$. Each initial-view image is repeated in time, encoded, and token-averaged; per-view vectors are concatenated into $h(o_0)$ and internally organized as view-condition tokens. Within the fixed task set, $g$ is mapped to a categorical identifier and learned task token rather than a free-form language embedding. Actions preserve their order and are projected into temporally corresponding tokens.

Trajectory features undergo spatial pooling and channel projection, followed by temporal, view, and spatial positional encodings. Full-window self-attention connects chunks and views; conditional cross-attention incorporates initial observations, task identity, and temporally corresponding actions. Learned-query pooling produces the scalar $D_\psi(z,c)$. Future trajectories are neither globally mean-pooled before processing nor scored as independent chunks.

\subsection{Conditional Wasserstein Alignment and Alternating Optimization}
\label{app:optimization}
Local generative supervision does not directly assess trajectory-wide dynamics. We learn a feature-distribution discrepancy under matched conditions and propagate its differentiable signal to the world model.

\paragraph{Critic and gradient penalty.}
We use a Wasserstein critic objective~\cite{arjovsky2017wasserstein}, corresponding to main-paper Eq.~(2):
\begin{equation}
\begin{aligned}
\mathcal L_D=&\frac1s\left(\mathbb E D_\psi(\hat z,c)
-\mathbb E D_\psi(z,c)\right)\\
&+\lambda_{\mathrm{gp}}\mathcal L_{\mathrm{gp}},
\qquad s=\sqrt d.
\end{aligned}
\end{equation}
Here $d$ counts feature elements per trajectory and $\lambda_{\mathrm{gp}}$ weights the penalty. A single $\alpha\sim\mathcal U[0,1)$ is sampled and broadcast per window:
\begin{equation}
\tilde z=\alpha z+(1-\alpha)\operatorname{sg}(\hat z),
\end{equation}
\begin{equation}
\mathcal L_{\mathrm{gp}}=
\mathbb E\left[(\|\nabla_{\tilde z}D_\psi(\tilde z,c)\|_2-1)^2\right].
\end{equation}
This follows WGAN-GP input-gradient regularization~\cite{gulrajani2017improved}. Conditions $c$ are held fixed and detached. Gradients are taken in standardized feature coordinates before spatial pooling. The unit-gradient target applies to $D_\psi$, not $D_\psi/s$.

\paragraph{World-model objective and weighting.}
Main-paper Eq.~(3) can be written as
\begin{equation}
\mathcal L_W=-\frac1s\mathbb E D_\psi(\hat z_\phi,c),
\qquad
\mathcal L_{\mathrm{world}}=
\mathcal L_{\mathrm{gen}}^{\mathrm{SR}}+\lambda_W\mathcal L_W.
\end{equation}
The notation $\hat z_\phi=S(F(\hat\tau_\phi))$ emphasizes parameter dependence. Critic, video-encoder, and VAE parameters are frozen during world-model updates, while their input derivatives are retained.

Let $g_F,g_W$ denote the flow-matching and alignment gradients. We calibrate the initial weight using sums of global norms over calibration batches. At update $n$, let $u=\min(n/T,1)$, $b=u\lambda_{\rm old}$ and $r=\|bg_W\|/\|g_F\|$. With numerical safeguard $\epsilon$, the weights are
\begin{equation}
\begin{aligned}
\lambda_0 &= 0.1\,\frac{\sum_i\|g_{F,i}\|}{\sum_i\|g_{W,i}\|},\\
\lambda_W^{(n)} &= b\min\{1,0.2u/\max(r,\epsilon)\},\\
\lambda_{\rm next} &= \operatorname{clip}\big(0.1/\max(E,\epsilon),\,
                 \lambda_{\rm old}/2,\,2\lambda_{\rm old}\big).
\end{aligned}
\end{equation}
At scheduled updates, initialize $E$ to $r/\max(b,\epsilon)$ and subsequently update it by exponential averaging with decay $\beta$. The new weight applies from the next step. The initial target ratio is 0.1 and the ratio cap reaches 0.2 after warm-in. Weight adaptation is detached.

\paragraph{Alternation and gradient paths.}
For each reference window, the world model rolls out the recorded actions using its own boundary frames. The corresponding recorded future provides the flow-matching target for each chunk, and the generated chunks together provide the input to the trajectory critic. We first update the critic with detached generated features and draw fresh interpolation coefficients when reusing a batch. We then freeze the updated critic, rescore the same generated trajectory, and update the world model with the combined loss $\mathcal L_{\mathrm{world}}$. Each chunk receives alignment gradients through the full-trajectory score along $\phi\rightarrow\hat O_{k+1}\rightarrow F(\hat\tau)\rightarrow D_\psi$. Detaching boundary frames truncates the gradient through later chunks conditioned on those frames. New windows and trajectories are sampled across iterations.

\subsection{Reward Model and Trajectory-Level Success Estimation}
\label{app:reward}
\paragraph{Model and supervision.}
The reward classifier takes a video window and the task instruction $g$ as inputs and estimates the probability that the instructed task has been completed. In simulation, its visual branch uses a shared VideoMAE-base backbone across the front and wrist views~\cite{tong2022videomae}. We average the final-layer visual tokens within each view and concatenate the two view representations. For real-robot tasks, the video comes from the fixed workspace camera described in Appendix~\ref{app:real-robot-protocol}. We train one classifier per simulator suite and one per real-robot task. Labels are constructed from execution outcomes and first-success times. Models are frozen thereafter.

Within each simulator suite or real-robot task, all policies and world models share the frozen classifier. Appendix~\ref{app:action-ranking-physical-events} defines the trajectory score, and Appendix~\ref{app:gd-policy-training} specifies threshold calibration and terminal rewards.

\subsection{Policy Optimization for VLAs and WAMs}
\label{app:policy}
To translate improved prediction into policy gains, we optimize action-transition likelihoods using terminal feedback in a frozen imagined environment, without backpropagating through the environment or reward computation.

\paragraph{Closed-loop rollout and advantages.}
For each group, the behavior policy $\pi_{\theta_{\mathrm{old}}}$ samples $G>1$ trajectories from the same initial observation and task instruction. At each decision, it samples an action chunk from the current observation. The actions are converted to the input representation of the frozen world model $M_{\phi^*}$, and the final generated frame supplies the next policy observation after preprocessing. We repeat this interaction until termination or the $J$-chunk horizon. The frozen reward model assigns a terminal reward as in Appendix~\ref{app:gd-policy-training}. With $R_i=R(\hat\tau_i,g)$,
\begin{equation}
\begin{aligned}
\bar R&=\frac1G\sum_i R_i,&
\sigma_R&=\sqrt{\frac1{G-1}\sum_i(R_i-\bar R)^2},\\
A_i&=\frac{R_i-\bar R}{\sigma_R+\eta}.&&
\end{aligned}
\end{equation}
Here $\eta>0$ is the main-paper advantage stabilizer. Advantages are held fixed and shared across valid chunks and transitions of each trajectory. Equal group rewards give zero advantages. This follows group-relative feedback in GRPO~\cite{shao2024deepseekmath}.

\paragraph{Stochastic action transitions.}
We define likelihoods through stochastic denoising transitions, following the flow-policy optimization approach of Flow-GRPO~\cite{liu2025flowgrpo}. Let $t_j>t_{j+1}$, $\Delta t_j=t_j-t_{j+1}$, $x^a_{t_j}$ be the action state, and $v_\theta$ the native predicted velocity. Define
\begin{equation}
x_0=x^a_{t_j}-t_jv_\theta,\qquad
x_1=x^a_{t_j}+(1-t_j)v_\theta,
\end{equation}
\begin{equation}
\begin{aligned}
\mu_\theta&=(1-t_{j+1})x_0+
\left(t_{j+1}-\frac{\nu_{t_j}^2\Delta t_j}{2t_j}\right)x_1,\\
\sigma_{\mathrm{tr},j}&=\sqrt{\Delta t_j}\,\nu_{t_j}.
\end{aligned}
\end{equation}
Here $\nu_{t_j}$ controls noise strength. WAM uses $\nu_{t_j}=\eta_{\mathrm{SDE}}t_j$. The VLA schedule and first-step convention appear in Table~\ref{tab:rl-actor-settings}. The noise coefficient $\eta_{\mathrm{SDE}}$ is distinct from the advantage stabilizer $\eta$. Transition schedules are specified in Appendix~E; only selected nonzero-variance transitions contribute Gaussian likelihoods.

\paragraph{Cached recomputation and reduction.}
Rollouts cache transition endpoints, conditions, indices, and behavior log-probabilities. Current parameters rescore the same transition:
\begin{equation}
\ell_\theta(i,k,j)=\sum_{p,r}\log\mathcal N
\left(x^a_{t_{j+1},p,r};\mu_{\theta,p,r},
\sigma_{\mathrm{tr},j}^2\right),
\end{equation}
Here $p$ indexes actions within a chunk and $r$ indexes valid action coordinates. Fixed conditions include observations, instructions, noise times, and required cached inputs. The ratio in main-paper Eq.~(5) is
\begin{equation}
\rho_{i,k,j}=\exp[\ell_\theta(i,k,j)-
\ell_{\theta_{\mathrm{old}}}(i,k,j)].
\end{equation}
Termination masks retain the terminal action and exclude later actions. A chunk is valid if any action in it is valid. Let $T_i$ be the number of valid original actions, $H$ the maximum action count, $m_{i,k}$ the chunk mask, and $\mathcal J_{i,k,j}$ the signed, per-transition clipped loss in main-paper Eq.~(5). The reduction is
\begin{equation}
\mathcal L_{\mathrm{policy}}=
\operatorname{mean}_{i,k,j}
\left[m_{i,k}\mathcal J_{i,k,j}\frac H{T_i}\right].
\end{equation}
The outer mean is over the fixed sampled tensor, rather than renormalizing by the valid-position count. Additional run-specific clipping settings are given in Appendix~E.

\paragraph{Gradient paths.}
WAM detaches cached joint latents before recomputation, but differentiates current network operations on those inputs. Action likelihoods can therefore update participating visual input projections, shared backbone parameters, and action modules. No separate video likelihood is introduced, and gradients do not traverse earlier cached transitions, the environment, or RM. VLA visual and language prefixes are frozen and evaluated without gradients; trainable action experts and associated projections receive gradients. Complete parameter scopes are given in Appendix~E.

\section{Reward-Model Validation and Trajectory Scoring}
\label{app:gd-data-training}
GroundedDream uses a reward model to estimate success in imagined rollouts and to supply terminal rewards for policy optimization. We first validate this signal on simulator videos with known outcomes, then describe how window-level predictions yield a trajectory score.

\subsection{Accuracy of Video-Based Success Judgments}
\label{app:reward-blind}
\label{app:gd-data}
A failure classified as a success can inflate the estimated success rate in imagination. To quantify both classification accuracy and the direction of this bias, we compare the frozen classifier's predictions with simulator outcomes on 500 trajectories from the official LIBERO-Spatial test set.

\begin{table}[t]
\centering
\caption{\textbf{Reward judgments on held-out simulator videos.} Rows give simulator outcomes and columns give reward-model predictions. Each trajectory contributes one observation.}
\label{tab:reward-blind}
\begin{tabular}{@{}lrrr@{}}
\toprule
& \multicolumn{2}{c}{Reward prediction} & \\
\cmidrule(lr){2-3}
Simulator outcome & Success & Failure & Total \\
\midrule
Success & 268 & 8   & 276 \\
Failure & 24  & 200 & 224 \\
\midrule
Total   & 292 & 208 & 500 \\
\bottomrule
\end{tabular}
\end{table}

The classifier correctly judges 468 of 500 trajectories (93.6\%; Table~\ref{tab:reward-blind}). It misses 8 of 276 successful executions and labels 24 of 224 failures as successes. The estimated success rate is consequently 58.4\%, compared with 55.2\% in the simulator. Thus, the classifier identifies task outcomes accurately on Spatial simulator videos, with a modest upward bias of 3.2 percentage points. Final policy performance is measured through execution: simulator experiments use the environment's native success signal, and real-robot trials use the predefined criteria in Appendix~\ref{app:real-robot-protocol}. The classifier's bias therefore does not enter the final success counts directly. Appendix~\ref{app:rl-reward-agreement} separately tests whether rewards during training agree with the consequences of replaying the same actions.

\paragraph{Simulator reward data.}
We train a separate classifier for each of Spatial, Long, Object, and Goal using the architecture in Appendix~\ref{app:reward}. For each suite, supervised fine-tuned (SFT) policies collect 2,500 complete trajectories. We use 2,000 to train the classifier and 500 to calibrate the success threshold. Window labels follow the simulator success signal and the first time the task succeeds. For Spatial, four executions per task and initial state supply training data and a fifth supplies calibration data. Once the classifier and threshold are fixed, we evaluate on the official test set for the corresponding suite, which contains 50 initial states for each of ten tasks. Table~\ref{tab:reward-blind} reports the Spatial results.

\paragraph{Real-robot reward data.}
We train a classifier separately for each of the four robot tasks. Of the 80 interaction trajectories collected by the SFT policy for each task, 50 train the classifier, 15 calibrate its threshold, and 15 test it. The same 80 trajectories also train the task-specific world model. The policy success rates in Appendix~\ref{app:real-robot-results} come from robot executions.

\subsection{Continuous Trajectory Scores and Terminal Rewards}
\label{app:gd-policy-training}
\label{app:action-ranking-physical-events}
The reward model estimates a success probability for each video window. Let $p_{i,(1)}$ and $p_{i,(2)}$ denote the largest and second-largest probabilities for trajectory $i$. For instruction $g_i$ and imagined trajectory $\hat\tau_i$, the continuous score $s_i=q(\hat\tau_i,g_i)$ in the main paper is
\begin{equation}
s_i=\tfrac12\bigl(p_{i,(1)}+p_{i,(2)}\bigr).
\label{eq:app-action-ranking-score}
\end{equation}
The selected windows need not be adjacent. For the 240-frame Spatial predictions, each window contains eight consecutive frames. Endpoints range from frame 64 to frame 240 at a stride of four, giving 45 windows per trajectory. The same rule gives 113 windows for each 512-frame Long prediction.

Action-sequence ranking in main-paper Table~1 uses $s_i$ directly. To estimate imagined success rates, we select a threshold by trajectory-level macro-F1 on the 500 calibration trajectories for each simulator suite and hold it fixed across policies and world models on that suite. In imagined RL, the terminal reward is 5 when the score reaches the threshold and 0 otherwise. Intermediate rewards are zero. Direct simulator RL uses the native LIBERO reward.

\section{Action-Conditioned World-Model Predictions and Physical Events}
Main-paper Tables~1 and~2 evaluate video quality, action-outcome ranking, and target localization. We extend this analysis in three ways: a regression control isolates the role of the conditional trajectory critic, manual event labels test whether improvements are visible during grasp and transport, and full LIBERO-Long trajectories test prediction over a longer horizon.

\subsection{Supplementary Ablation of Trajectory Alignment}
\label{app:paired-feature-regression}
\label{app:gd-world-training}
\label{app:gd-baselines}
\paragraph{Motivation.}
Main-paper Table~2 removes trajectory alignment, scores chunks independently, removes action conditioning from the critic, and replaces the predictive representation. One question remains: given the same V-JEPA~2 features, does direct regression to the paired reference features recover the gains of a learned critic?

\paragraph{Ablation.}
We add paired feature regression between the no-alignment control and Full PDA. Regression and Full PDA share the training data, two-view inputs, frozen encoder, autoregressive rollout, flow-matching supervision, and world-model update budget. Regression minimizes the elementwise mean squared error between generated and reference trajectory features. Full PDA instead trains an action-conditioned critic to score predictive features jointly across consecutive chunks.

\begin{table*}[t]
\centering
\caption{\textbf{Paired feature regression on LIBERO-Spatial.} Video metrics are grouped by view. AUROC and front-view object-center error are reported once per model. Bold and underlining indicate the best and second-best values in each column, respectively.}
\label{tab:paired-regression}
\footnotesize
\setlength{\tabcolsep}{1.5pt}
\begin{tabular*}{\textwidth}{@{\extracolsep{\fill}}l*{12}{c}@{}}
\toprule
\multirow{2}{*}{Method} & \multicolumn{5}{c}{Front View} & \multicolumn{5}{c}{Wrist View}
& \multirow{2}{*}{AUROC$\uparrow$} & \multirow{2}{*}{\shortstack{Object-Center\\Error (px)$\downarrow$}}\\
\cmidrule(lr){2-6}\cmidrule(lr){7-11}
& LPIPS$\downarrow$ & FID$\downarrow$ & FVD$\downarrow$ & SSIM$\uparrow$ & PSNR$\uparrow$
& LPIPS$\downarrow$ & FID$\downarrow$ & FVD$\downarrow$ & SSIM$\uparrow$ & PSNR$\uparrow$ & &\\
\midrule
w/o Trajectory Alignment & 0.072 & 19.50 & 42.59 & 0.840 & 22.79 & \underline{0.205} & \underline{33.37} & 98.88 & \underline{0.707} & 17.58 & 0.791 & 8.06\\
Paired Feature Regression & \underline{0.069} & \underline{10.68} & \underline{39.11} & \underline{0.852} & \underline{23.24} & 0.214 & 35.04 & \underline{90.46} & 0.704 & \underline{18.26} & \underline{0.813} & \underline{7.23}\\
Full PDA & \textbf{0.061} & \textbf{8.96} & \textbf{35.77} & \textbf{0.862} & \textbf{23.94} & \textbf{0.198} & \textbf{30.80} & \textbf{85.64} & \textbf{0.710} & \textbf{18.69} & \textbf{0.849} & \textbf{5.47}\\
\bottomrule
\end{tabular*}
\end{table*}

Relative to removing alignment, paired regression raises AUROC from 0.791 to 0.813 and reduces target-center error from 8.06 to 7.23 pixels (Table~\ref{tab:paired-regression}). Predictive features thus provide useful supervision even with a direct regression target. Using the conditional trajectory critic in the same feature space further increases AUROC to 0.849 and reduces center error to 5.47 pixels. Full PDA also improves all five video metrics in both views over regression. Together with main-paper Table~2, these results identify an additional benefit of learning how to compare action consequences in predictive feature space.

\paragraph{World-model data and comparison conditions.}
For each suite--backbone combination evaluated on Spatial, Long, Object, or Goal, we execute the corresponding SFT policy (Cosmos~3 Edge WAM, $\pi_0$, or $\pi_{0.5}$) to collect 2,500 trajectories. We split complete episodes into 2,000 for world-model training and 500 for validation. Final evaluation uses each suite's official test set of 50 initial states per task and 500 per suite. Training windows and PDA feature-normalization statistics come from the training split.

Within a suite and policy backbone, all methods share the training and validation splits. Unaligned is the base world-model reference, and Ctrl-World is an action-conditioned baseline. Self Forcing and GroundedDream use the same starting checkpoint and training data. The alignment ablations in main-paper Table~2 and supplementary Table~\ref{tab:paired-regression} retain self-rollout training and flow matching.

\paragraph{Training windows and optimization.}
The simulator's front and wrist cameras and robot controller operate at 20\,Hz. We resize each view to $256\times256$ and concatenate them horizontally, front view first, to obtain a $256\times512$ observation. Each chunk covers 16 consecutive actions and 16 future observations (0.8\,s); a shared boundary frame is counted once. We sample a task uniformly, then extract $K$ consecutive chunks from one of its training trajectories. The default $K=4$ covers 64 actions and future observations (3.2\,s). The $K=2,4,8$ settings span 1.6, 3.2, and 6.4\,s during training and use the same 12\,s Spatial test horizon.

The default Spatial run uses eight NVIDIA H100 GPUs, with one four-chunk window per GPU and a global batch of eight. Learning rates are $2\times10^{-6}$ for the main parameters and $10^{-5}$ for the action interface. The generator takes four Euler steps on the grid $1,0.75,0.5,0.25,0$ with the gradient truncation described in Appendices~B.2--B.4. Training takes approximately one to two days on this hardware.

The reference and generated branches encode simulator RGB observations and VAE-decoded observations, respectively, using a shared frozen V-JEPA~2 encoder and channel-wise normalization. We use 64 training windows to estimate the normalization statistics and another 64 to calibrate the alignment weight. The critic uses Adam with learning rate $10^{-4}$, $(\beta_1,\beta_2)=(0,0.9)$, and gradient penalty coefficient 10. After 400 critic warm-up steps, each world-model update is paired with two critic updates. The target alignment-to-flow-matching gradient norm ratio is 0.1. We phase in alignment over the first 200 updates and adjust its weight every ten updates using the moving average of the gradient norm ratio. The weighted alignment gradient norm is capped at 0.2 times the flow-matching gradient norm.

\paragraph{Matched-action evaluation.}
For each held-out reference execution, all world models receive the same initial observations, task instruction, and recorded action sequence. Each model generates its rollout autoregressively, conditioning each subsequent chunk on its own preceding prediction. We compare each generated rollout with the simulator trajectory recorded for that same action sequence to assess how faithfully the model predicts the observed consequences of the executed actions. On Spatial, each rollout spans 15 chunks (240 future frames).

\paragraph{Evaluation metrics.}
\label{app:video-prediction-localization}
For frame-level metrics, we pair each prediction with the simulator observation at the same time step and exclude the conditioning frame.

\emph{Video quality.} We compute PSNR, SSIM, and LPIPS per frame, then average within trajectories and across the 500 test trajectories. SSIM uses RGB in $[0,1]$, an $11\times11$ Gaussian window with $\sigma=1.5$, and population covariance. LPIPS uses AlexNet v0.1. FID uses 2,048-dimensional Inception features per frame; FVD uses one 400-dimensional I3D feature per complete video. We report each view separately and use their arithmetic mean when aggregating views.

\emph{Target localization.} We initialize the target tracker in the shared $256\times256$ front-view frame and track the object separately in generated and reference videos. The frame-level error is
\begin{equation}
e_{i,t}=\lVert\hat{\mathbf c}_{i,t}-\mathbf c_{i,t}\rVert_2.
\label{eq:app-object-center-error}
\end{equation}
We average within trajectories and then over test trajectories.

\emph{Action-outcome ranking.} We pair the continuous generated-video score $s_i$ with the simulator outcome for the same actions. The 500 Spatial world-model test trajectories contain 284 successes and 216 failures. For each success--failure pair, we assign 1 if the successful sequence scores higher, 0.5 for a tie, and 0 otherwise. AUROC is the mean over pairs. These counts describe the reference executions for prediction evaluation; Table~\ref{tab:reward-blind} describes the executions used to test the reward classifier.

\subsection{Agreement on Grasp and Transport Events}
\label{sec:physical_event_evaluation}
Main-paper Fig.~1(a) illustrates how similar images can depict different grasp outcomes. On two bowl-placement tasks, we manually annotate grasp and transport outcomes in the generated videos and their matched simulator executions, then report event-level agreement with the simulator labels. We evaluate transport agreement on executions in which the simulator shows a successful grasp. The manual annotations are independent of the reward classifier used to compute AUROC.

\begin{table}[t]
    \centering
    \caption{Observable criteria for physical-event annotation in placement tasks. Each event is labeled separately in the reference and generated videos.}
    \label{tab:app-physical-event-criteria}
    \begin{tabular}{@{}p{0.20\linewidth}p{0.73\linewidth}@{}}
        \toprule
        \textbf{Event} & \textbf{Criterion for \texttt{yes}} \\
        \midrule
        Grasping &
        The target object lifts off its original support surface with the gripper. \\
        Carrying &
        After grasping the object, the gripper retains it throughout transport until release. \\
        Placement &
        After release, the designated plate supports the object, and subsequent frames show that it remains stable. \\
        \bottomrule
    \end{tabular}
\end{table}

Table~\ref{tab:app-physical-event-criteria} distinguishes lifting the target, retaining it during transport, and achieving stable placement. Annotators label each event independently as \texttt{yes}, \texttt{no}, or \texttt{unknown} in the generated and reference videos. A lift followed by a drop counts as a successful grasp and failed transport. A clearly missed grasp leaves the later stages unreached. When occlusion or insufficient detail prevents a judgment, only the affected event receives an unknown label. For example, a video ending at release or first contact with the plate provides insufficient evidence of stable placement.

\begin{table}[t]
    \centering
    \caption{\textbf{Agreement with simulator reference videos on physical
    events.} Each cell reports the number of matching event labels over
    the number of evaluated trajectories, followed by the agreement rate.
    Transport is evaluated conditional on a successful grasp in the
    reference video.}
    \label{tab:physical_event_agreement}
    \begin{tabular}{@{}lcc@{}}
        \toprule
        Event & Unaligned & Full PDA \\
        \midrule
        \multicolumn{3}{@{}l}{\emph{Place Ramekin-Side Bowl on Plate}} \\
        Grasp     & 20/30 (66.7\%) & \textbf{23/30 (76.7\%)} \\
        Transport & 10/18 (55.6\%) & \textbf{14/18 (77.8\%)} \\
        \midrule
        \multicolumn{3}{@{}l}{\emph{Place Center Bowl on Plate}} \\
        Grasp     & 29/30 (96.7\%) & \textbf{30/30 (100.0\%)} \\
        Transport & 24/27 (88.9\%) & \textbf{26/27 (96.3\%)} \\
        \bottomrule
    \end{tabular}
\end{table}

Full PDA agrees more often with the simulator on both events in both tasks (Table~\ref{tab:physical_event_agreement}). For the ramekin-side bowl, Full PDA correctly predicts the transport outcome in 14 of the 18 trajectories in which the reference gripper lifts the target, compared with 10 for Unaligned. For the center bowl, both models already reproduce nearly all grasps, yet transport agreement increases from 24/27 to 26/27. These judgments are independent of the reward classifier and connect the gains in the main paper to visible consequences of the actions.

Grasp agreement uses all 30 trajectories per task. Transport agreement uses the same reference-defined subset for both models. If a reference grasp succeeds but a generated grasp fails, the trajectory remains in the transport denominator with a generated transport label of \texttt{no}.

\subsection{Long-Horizon Prediction on LIBERO-Long}
\label{app:long-prediction}
Main-paper Table~1 reports LIBERO-Spatial predictions and refers to supplementary evaluation on other datasets. We evaluate the same four models on full LIBERO-Long trajectories to test whether the findings extend to longer tasks. Long follows the protocol in Appendix~\ref{app:gd-world-training}, with 500 official test initial states across all ten tasks. Each model follows the same recorded actions for 32 consecutive chunks, conditioning each chunk on the final predicted frame of the preceding one. We score all 512 future frames (25.6\,s).

\begin{table*}[t]
\centering
\caption{\textbf{Prediction on complete LIBERO-Long trajectories.} AUROC measures action-sequence ranking across trajectories. Object-center error is measured in the front view. Bold and underlining indicate the best and second-best values in each column, respectively.}
\label{tab:long-prediction}
\footnotesize
\setlength{\tabcolsep}{1.5pt}
\begin{tabular*}{\textwidth}{@{\extracolsep{\fill}}l*{12}{c}@{}}
\toprule
\multirow{2}{*}{Method} & \multicolumn{5}{c}{Front View} & \multicolumn{5}{c}{Wrist View}
& \multirow{2}{*}{AUROC$\uparrow$} & \multirow{2}{*}{\shortstack{Object-Center\\Error (px)$\downarrow$}}\\
\cmidrule(lr){2-6}\cmidrule(lr){7-11}
& LPIPS$\downarrow$ & FID$\downarrow$ & FVD$\downarrow$ & SSIM$\uparrow$ & PSNR$\uparrow$
& LPIPS$\downarrow$ & FID$\downarrow$ & FVD$\downarrow$ & SSIM$\uparrow$ & PSNR$\uparrow$ & &\\
\midrule
Unaligned world model & 0.534 & 109.03 & 586.53 & 0.670 & 17.24 & 0.478 & 79.66 & 447.98 & \underline{0.653} & \underline{16.18} & \underline{0.593} & \underline{20.84}\\
Ctrl-World & 0.487 & \underline{38.26} & \underline{119.67} & \underline{0.746} & \underline{18.07} & \underline{0.422} & \underline{56.16} & \underline{141.88} & 0.630 & 15.79 & 0.590 & 25.55\\
Self Forcing (DMD) & \underline{0.362} & 90.44 & 330.95 & 0.580 & 17.58 & 0.551 & 67.68 & 266.93 & 0.534 & 15.26 & 0.584 & 21.12\\
GroundedDream & \textbf{0.146} & \textbf{13.29} & \textbf{59.92} & \textbf{0.794} & \textbf{19.11} & \textbf{0.323} & \textbf{27.07} & \textbf{71.18} & \textbf{0.671} & \textbf{17.23} & \textbf{0.665} & \textbf{14.27}\\
\bottomrule
\end{tabular*}
\end{table*}

GroundedDream achieves the best value for all five video metrics in both views, the highest AUROC, and the lowest target-center error (Table~\ref{tab:long-prediction}). Relative to Ctrl-World, the strongest FVD baseline, front-view FVD decreases from 119.67 to 59.92 and wrist-view FVD from 141.88 to 71.18. Relative to Unaligned, the strongest ranking and localization baseline, AUROC increases from 0.593 to 0.665 and center error decreases from 20.84 to 14.27 pixels. The improvements in video prediction, action-outcome ranking, and localization observed on Spatial thus extend to the longer autoregressive rollouts on Long.

\section{Closed-Loop Policy Evaluation and Optimization}
\label{app:policy-evaluation-optimization}
The main paper uses world models both to evaluate fixed policies and to train new ones. We first compare closed-loop success estimates with simulator outcomes for the same policy. We then optimize policies in each world model and evaluate the resulting policies through execution in the simulator or on the robot.

\subsection{Policy Initialization and Closed-Loop Training}
\label{app:rl-settings-results}
We evaluate a Cosmos~3 Edge WAM and the $\pi_0$ and $\pi_{0.5}$ VLAs. For simulator SFT, Spatial and Long each provide 50 successful official demonstrations for each of ten tasks. We randomly select one demonstration per participating task and use the same demonstration across backbones. Within a task and backbone, RL runs start from the same SFT policy and are compared at matched update rounds with fixed reward models and optimization settings. The WAM policy and the action-conditioned world model are initialized separately. The world model and reward model remain frozen during policy training.

The WAM stove-and-moka-pot task in main-paper Figs.~4 and~5 uses the Long world model described in Appendix~\ref{app:long-prediction}. For the Object and Goal results in main-paper Fig.~6, the SFT policy for each suite--VLA combination collects the training data, and its separately trained world-model checkpoint serves as the imagined environment.

\paragraph{Closed-loop interaction and update budget.}
The policy predicts an action chunk from the current observation and instruction. The world model predicts the resulting observations and returns them for the next decision. At termination, the reward in Appendix~\ref{app:gd-policy-training} determines the group-relative advantage in Flow-GRPO. For the two LIBERO-Spatial WAM bowl-placement experiments, each world model retains 64 groups containing both successful and failed trajectories per round, for a total of 512 trajectories used for policy updates. Groups whose trajectories all succeed or all fail have zero group-relative advantages and are excluded from updates. The compared world models use the same total number of training rounds and optimizer steps.

\begin{table*}[t]
\centering
\caption{\textbf{Actor-only GRPO settings.} The WAM column covers joint Video + Action updates on the two bowl-placement tasks. The $\pi_{0.5}$ column describes the actor configuration. Batch sizes refer to optimizer inputs. The VLA noise schedule is $\nu_{t_j}=\eta_{\mathrm{SDE}}\sqrt{t_j/(1-d_j)}$, with $d_j=t_{j+1}$ on the first step and $d_j=t_j$ thereafter.}
\label{tab:rl-actor-settings}
\begin{tabular}{@{}lll@{}}
\toprule
Setting & Cosmos~3 Edge WAM & $\pi_{0.5}$ \\
\midrule
Group size / optimizer epochs & 8 / 1 & 8 / 1 \\
Actions per policy query & 16 & 5 \\
Denoising time grid & $1,\,3/4,\,1/2,\,1/4,\,0$ & $1,\,2/3,\,1/3,\,0$ \\
Stochastic transitions & First three & One selected per sampling batch \\
Optimizer updates per outer round & 3 & 12 \\
Actor batch / microbatch & 2560 / 8 & 2048 / 32 \\
Learning rate & $10^{-5}$ & $5\times10^{-6}$ \\
PPO ratio clipping / gradient norm cap & $[0.8,1.2]$ / 1.0 & $[0.8,1.2]$ / 1.0 \\
Additional log-ratio clipping & $[-20,20]$ & Disabled \\
\bottomrule
\end{tabular}
\end{table*}

Table~\ref{tab:rl-actor-settings} specifies the action horizon, stochastic transitions, and optimizer settings for each backbone. These settings are fixed when comparing world models within a backbone. Both configurations use Adam with $(\beta_1,\beta_2)=(0.9,0.95)$, $\epsilon=10^{-8}$, and weight decay 0.01. Within-group normalization uses the sample standard deviation plus $10^{-6}$. KL and entropy coefficients are zero and the noise parameter is 0.5. WAM uses transition order $(2,1,0)$ with learning-rate multipliers $(1,1,0.5)$. The VLA shares the sampled transition within a batch; its noise schedule is given in the table caption. Both recompute action log-likelihoods on cached transitions.

\paragraph{Parameter update scopes.}
Main-paper Fig.~5 compares three WAM settings. Action-only updates train the action input and output projections and the action modality embedding. Video-only updates train the video latent input and output projections. Joint updates train both interfaces and the shared generative parameters participating in action prediction, including the time embedding, generation expert, and associated key normalization. All three optimize the log-likelihood of action-denoising transitions. For $\pi_{0.5}$, updates cover the action expert's attention, MLP and normalization layers, action input and output projections, and time MLP projections. The visual and language prefix remains frozen.

\subsection{Closed-Loop Policy Evaluation and Initial Placement Shifts}
\label{app:closed-loop-evaluation}
\label{app:vla-closed-loop}
\paragraph{Evaluation from official test states.}
Main-paper Fig.~4 compares imagined and simulator success rates for WAM checkpoints. We extend this comparison to checkpoints of $\pi_0$ and $\pi_{0.5}$. For each task and checkpoint, simulator and world-model rollouts begin from matched official test states (50 per task, 500 per suite). The policy then acts on observations returned by the respective environment, so subsequent actions can diverge.

\begin{figure*}[t]
    \centering
    \newcommand{\vlaclosedlooppanel}[1]{%
        \setlength{\unitlength}{1pt}%
        \resizebox{0.79\textwidth}{!}{%
            \begin{picture}(264,152)
                \put(0,0){\includegraphics[width=264pt]{#1}}
                \put(181.5,52.7){\color{white}\rule{47pt}{5.6pt}}
                \put(181.5,40.6){\color{white}\rule{47pt}{5.4pt}}
                \put(181.8,54){\sffamily\fontsize{4.6}{5.5}\selectfont Unaligned world model}
                \put(181.8,41.2){\sffamily\fontsize{4.6}{5.5}\selectfont Ctrl-World}
            \end{picture}%
        }%
    }
    \vlaclosedlooppanel{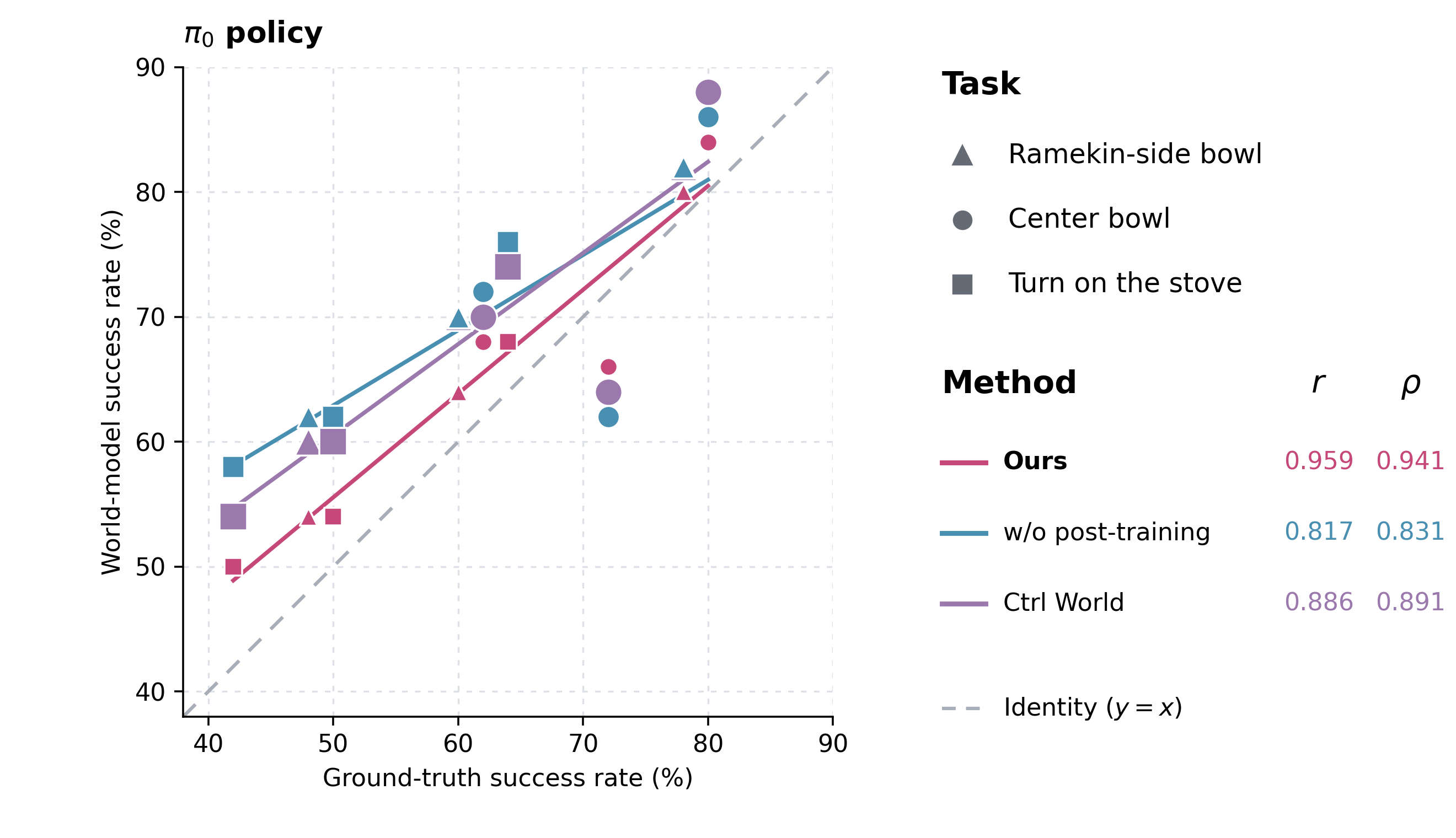}\par
    \smallskip
    \vlaclosedlooppanel{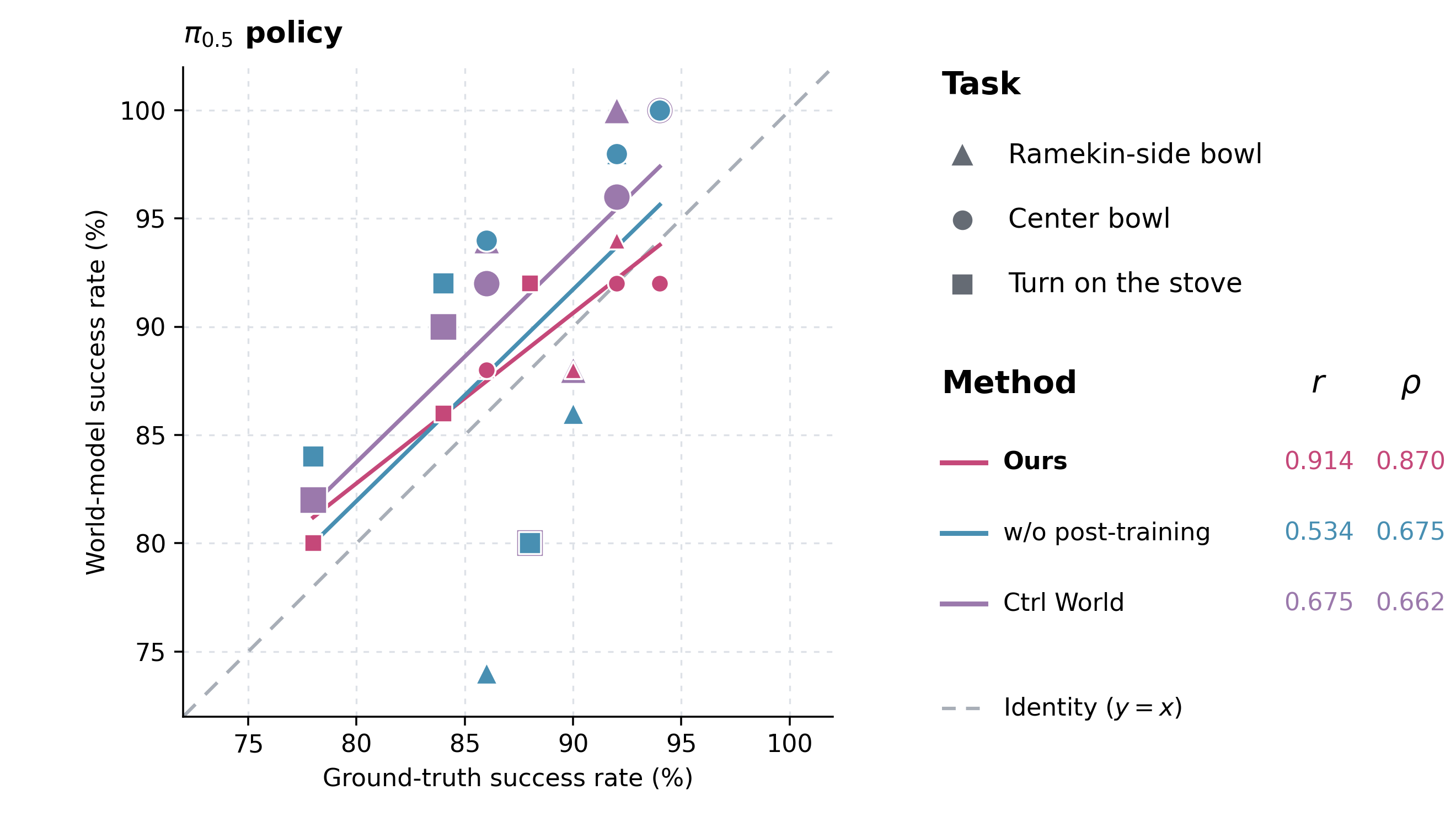}
    \caption{\textbf{Closed-loop policy evaluation for $\pi_0$ (top)
    and $\pi_{0.5}$ (bottom).} Each point pairs a simulator success
    rate with the world-model estimate for the same task and checkpoint.
    Marker shapes identify tasks and colors identify world models.
    The dashed line denotes equal success rates. Each legend reports
    Pearson's $r$ and Spearman's $\rho$ over all task--checkpoint pairs
    in that panel.}
    \label{fig:vla-closed-loop}
\end{figure*}

In Fig.~\ref{fig:vla-closed-loop}, each point represents one task--checkpoint pair. The horizontal axis gives simulator success and the vertical axis gives estimated success in imagination. Points above the identity line indicate overestimation. Within each panel, we pool the plotted pairs to calculate Pearson's $r$ and Spearman's $\rho$. The two bowl-placement tasks last 12\,s. WAM rollouts on the Long stove-and-moka-pot sequence last approximately 26\,s. The separate VLA stove task succeeds when the stove is turned on.

For $\pi_0$, Full PDA reaches $r=0.959$ and $\rho=0.941$, compared with $0.817/0.831$ for Unaligned and $0.886/0.891$ for Ctrl-World. For $\pi_{0.5}$, the corresponding values are $0.914/0.870$, $0.534/0.675$, and $0.675/0.662$. For example, one $\pi_0$ stove checkpoint succeeds in 42\% of simulator trials. Unaligned, Ctrl-World, and Full PDA estimate 58\%, 54\%, and 50\%, respectively, reducing overestimation from 16 points for Unaligned to 8 points for Full PDA. Across the evaluated task--checkpoint pairs, the stronger linear and rank correlations extend the main-paper finding to both VLA backbones.

\paragraph{Initial object-position shifts.}
\label{app:initial-placement-generalization}
The preceding comparison assesses differences between policies at standard placements. To test whether a world model also captures a change in performance for a \emph{fixed} policy, we construct 100 additional initial states and vary only the tabletop position of the target black bowl in two placement tasks. Bowl orientation, other objects, robot state, cameras, and instruction remain fixed. ID denotes new positions within the world-model post-training placement range; OOD positions fall outside it.

\begin{table}[t]
    \centering
    \caption{\textbf{Initial positions of the target bowl.}
    Coordinates are in meters in the simulator scene frame. For
    Task~2 under ID placements, $x$ is fixed and only $y$ varies.}
    \label{tab:initial-placement-ranges}
    \setlength{\tabcolsep}{4pt}
    \begin{tabular*}{\columnwidth}{@{\extracolsep{\fill}}llccr@{}}
        \toprule
        \textbf{Task} & \textbf{Split} & \textbf{$x$ (m)}
        & \textbf{$y$ (m)} & \textbf{States} \\
        \midrule
        1 & ID  & $[-0.19,-0.17]$ & $[0.31,0.33]$     & 25 \\
        1 & OOD & $[-0.28,-0.25]$ & $[0.30,0.33]$     & 25 \\
        2 & ID  & $-0.075$        & $[-0.01,0.01]$    & 25 \\
        2 & OOD & $[-0.18,-0.14]$ & $[-0.015,0.015]$  & 25 \\
        \bottomrule
    \end{tabular*}
\end{table}

In both tasks, OOD placements shift the target primarily in the negative $x$ direction (Table~\ref{tab:initial-placement-ranges}). We define the sampling regions in advance from the source scene, position sampler, and initial-state pool; eligible targets are visible and reachable, and initial scenes are collision-free. Selection is independent of policy success. We save the simulator state, scene XML, and initial images from both views. A fixed SFT policy runs once from each state in the simulator, Unaligned, and Full PDA, giving 300 closed-loop rollouts. Each rollout allows 240 control steps at 20\,Hz, with 16 actions per query. Simulator success requires meeting the task criterion within this horizon. Both world models use the same frozen reward classifier and precalibrated threshold.

\begin{table*}[t]
    \centering
    \caption{\textbf{Policy evaluation under initial object-position
    shifts.} Each row uses the same 25 initial states across
    environments. Absolute error is measured against the simulator
    success rate in percentage points (pp).}
    \label{tab:initial-placement-results}
    \begin{tabular*}{\textwidth}{@{\extracolsep{\fill}}llcccccc@{}}
        \toprule
        & & \multicolumn{2}{c}{\textbf{Simulator}}
          & \multicolumn{2}{c}{\textbf{Estimated success rate}}
          & \multicolumn{2}{c}{\textbf{Absolute error (pp)}} \\
        \cmidrule(lr){3-4}
        \cmidrule(lr){5-6}
        \cmidrule(l){7-8}
        \textbf{Task} & \textbf{Split}
          & \textbf{Successes/trials} & \textbf{Success rate}
          & \textbf{Unaligned} & \textbf{Full PDA}
          & \textbf{Unaligned} & \textbf{Full PDA} \\
        \midrule
        1 & ID  & 17/25 & 68\% &  60\% & 64\% &  8 & \textbf{4} \\
        1 & OOD &  0/25 &  0\% &  36\% &  4\% & 36 & \textbf{4} \\
        2 & ID  & 21/25 & 84\% & 100\% & 92\% & 16 & \textbf{8} \\
        2 & OOD & 17/25 & 68\% &  88\% & 76\% & 20 & \textbf{8} \\
        \bottomrule
    \end{tabular*}
\end{table*}

The two tasks respond differently to the shift (Table~\ref{tab:initial-placement-results}). Simulator success drops by 68 percentage points on Task~1; Unaligned predicts a 24-point drop and Full PDA a 60-point drop. On Task~2, the actual drop is 16 points; the two models predict drops of 12 and 16 points. Averaged equally over tasks, Unaligned's absolute error increases from 12 points under ID placements to 28 under OOD placements. Full PDA's error is 6 points under both conditions. Full PDA thus estimates both success rates and their sensitivity to placement more accurately in these two experiments.

\subsection{Reward Agreement during Optimization and Performance across Seeds}
\label{app:optimized-policy-evaluation}
Main-paper Figs.~5 and~6 compare policies trained in different world models. For WAM, joint video and action updates raise mean simulator success across the three evaluated tasks from 53.3\% for SFT to 90.0\% for Full PDA RL. VLA results are reported by suite on Spatial, Object, and Goal. We examine whether rewards assigned during training track actual action outcomes and whether final gains persist across independent training seeds.

\paragraph{Simulator action replay.}
\label{app:rl-reward-agreement}
Appendix~\ref{app:reward-blind} tests the classifier on simulator videos. During imagined RL, the classifier receives generated videos. To test whether the rewards assigned in imagination agree with outcomes of the same actions, we sample WAM at RL rounds 0, 25, and 50 on the ramekin-side bowl task using seed 42. Each method contributes 64 groups of eight trajectories per round, for 4,608 trajectories overall. We record the policy's actions in its world model, restore the simulator to the same initial state, and replay those actions. The native simulator label supplies the reference for macro-F1, averaged over the success and failure classes.

\begin{table}[t]
\centering
\caption{\textbf{Reward agreement during RL.} Trajectory-level macro-F1
(\%) against simulator action replay. Each entry uses 512 trajectories.}
\label{tab:rl-reward-agreement}
\renewcommand{\arraystretch}{1.14}
\begin{tabular*}{\columnwidth}{@{\extracolsep{\fill}}lrrr@{}}
\toprule
RL round & Unaligned & Ctrl-World & Full PDA \\
\midrule
0  & 63.36 & 64.52 & \textcolor[HTML]{49365E}{\textbf{69.99}} \\
25 & 68.41 & 69.42 & \textcolor[HTML]{49365E}{\textbf{73.18}} \\
50 & 80.33 & 80.52 & \textcolor[HTML]{49365E}{\textbf{86.17}} \\
\bottomrule
\end{tabular*}
\end{table}

Table~\ref{tab:rl-reward-agreement} reports how closely the rewards that policies receive during closed-loop training agree with simulator outcomes. At round 50, reward agreement reaches 86.17\% macro-F1 for Full PDA and 80.33\% for the Unaligned world model. Each score is computed on trajectories collected by the policy through interaction with its own world model.

\paragraph{Independent training seeds.}
\label{app:rl-seed-results}
To test whether gains recur across training runs, we initialize both environments from the same WAM SFT policy and train with joint video and action updates using seeds 42, 1, and 2 on the two bowl-placement tasks. We evaluate each final policy in the simulator from the same 50 official test states per task using the native success criterion.

\begin{table*}[t]
\centering
\caption{\textbf{Final simulator success across independent RL seeds.}
Seed columns give successes out of 50 trials and the corresponding success
rate. Means weight the three seeds equally. SD is the sample standard
deviation in percentage points, and $s^2$ is the sample variance in
squared percentage points.}
\label{tab:rl-seed-results}
\renewcommand{\arraystretch}{1.13}
\begin{tabular*}{\textwidth}{@{\extracolsep{\fill}}lcccrr@{}}
\toprule
& \multicolumn{3}{c}{Training seed} & \multicolumn{2}{c}{Across seeds} \\
\cmidrule(lr){2-4}\cmidrule(l){5-6}
RL world model & 42 & 1 & 2 & Mean (\%) $\pm$ SD & $s^2$ \\
\midrule
\multicolumn{6}{@{}l}{\emph{Place Ramekin-Side Bowl on Plate}} \\
Unaligned & 36/50 (72\%) & 38/50 (76\%) & 35/50 (70\%)
          & $72.7 \pm 3.1$ & 9.33 \\
Full PDA  & 44/50 (88\%) & 45/50 (90\%) & 43/50 (86\%)
          & \textcolor[HTML]{49365E}{$\mathbf{88.0} \pm 2.0$} & 4.00 \\
\midrule
\multicolumn{6}{@{}l}{\emph{Place Center Bowl on Plate}} \\
Unaligned & 37/50 (74\%) & 39/50 (78\%) & 40/50 (80\%)
          & $77.3 \pm 3.1$ & 9.33 \\
Full PDA  & 50/50 (100\%) & 49/50 (98\%) & 46/50 (92\%)
          & \textcolor[HTML]{49365E}{$\mathbf{96.7} \pm 4.2$} & 17.33 \\
\bottomrule
\end{tabular*}
\end{table*}

Full PDA improves final simulator success in all six task--seed comparisons (Table~\ref{tab:rl-seed-results}). Mean gains are 15.3 points for the ramekin-side bowl and 19.3 for the center bowl, with greater variation in the latter task. The direction of the main-paper gain is consistent across independent runs on both tasks. Means and sample standard deviations are computed over training seeds.

\subsection{Contribution of Trajectory Alignment to Policy Performance}
\label{app:downstream-rl-alignment-ablation}
Main-paper Table~2 and Appendix~\ref{app:paired-feature-regression} examine alignment during open-loop prediction. To test whether the same component improves the policies ultimately executed, we train policies in world models with alignment removed, with independent chunk-wise alignment, and with joint alignment across chunks. All variants use the same task, initial Cosmos~3 Edge WAM policy, and downstream update settings. This component ablation uses seed 42 and the same 50 official simulator test states per task.

\begin{table*}[t]
    \centering
    \caption{\textbf{Effect of world-model alignment on final policy
    success.} Each entry gives simulator successes out of 50 trials
    and the corresponding success rate. All RL results use training
    seed 42. SFT denotes the initial policy before RL.}
    \label{tab:downstream-rl-alignment}
    \begin{tabular*}{\textwidth}{@{\extracolsep{\fill}}lcc@{}}
        \toprule
        \textbf{Policy or RL training environment}
        & \textbf{Place Ramekin-Side Bowl on Plate}
        & \textbf{Place Center Bowl on Plate} \\
        \midrule
        SFT policy                 & 21/50 (42\%) & 37/50 (74\%) \\
        Unaligned world model      & 36/50 (72\%) & 37/50 (74\%) \\
        \midrule
        w/o Trajectory Alignment   & 37/50 (74\%) & 38/50 (76\%) \\
        w/ Chunk-wise Alignment    & 40/50 (80\%) & 43/50 (86\%) \\
        Full PDA                   & \textbf{44/50 (88\%)}
                                   & \textbf{50/50 (100\%)} \\
        \bottomrule
    \end{tabular*}
\end{table*}

Relative to removing alignment, chunk-wise alignment improves success by 6 and 10 percentage points on the two tasks. Joint alignment yields a further four and seven successful placements over chunk-wise alignment. Full PDA improves over the no-alignment variant by 14 and 24 points (Table~\ref{tab:downstream-rl-alignment}). Thus, the benefit of supervising consecutive chunks jointly extends from action-outcome prediction to actual simulator performance on these tasks. This component comparison uses seed 42; Table~\ref{tab:rl-seed-results} compares Full PDA with Unaligned across seeds.

\subsection{Policy Execution on the Real Robot}
\label{app:real-robot-results}
\label{app:real-robot-protocol}
Main-paper Fig.~7 reports mean policy gains over four real-robot tasks. To show how those gains are distributed, we evaluate the initial SFT policy, the policy optimized in the Unaligned world model, and the policy optimized in Full PDA on the same platform and at the same predefined placements.

\begin{figure}[t]
    \centering
    \includegraphics[width=0.78\linewidth]{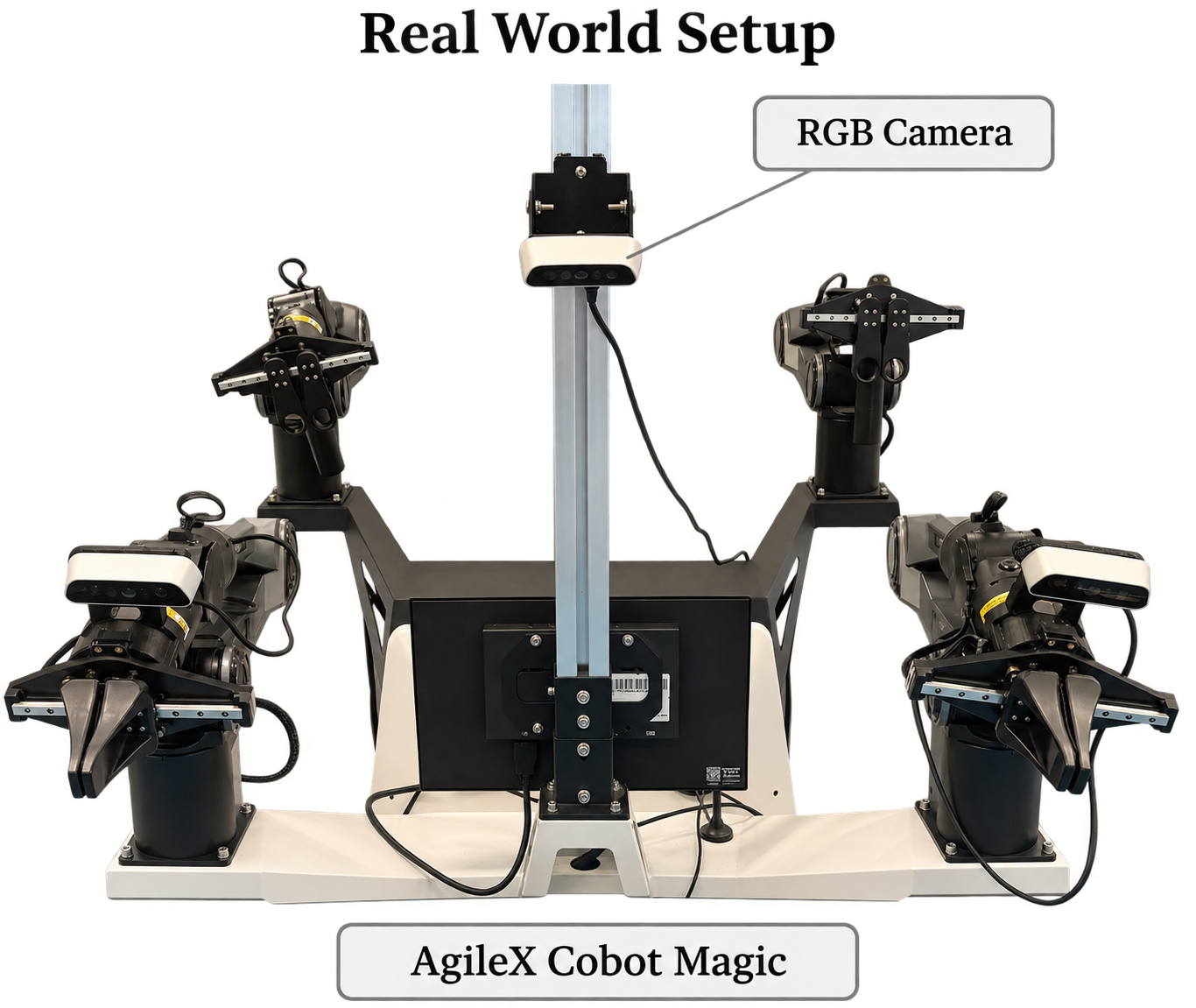}
    \caption{\textbf{Real-robot platform and camera setup.}
    A fixed RGB camera is mounted between the two robot arms above the
    workspace.}
    \label{fig:real_robot_setup}
\end{figure}

\paragraph{Platform and training data.}
All three policies use the Cosmos~3 Edge Unified WAM on the AgileX Cobot Magic dual-arm platform. A fixed RGB camera is mounted between the arms above the workspace (Fig.~\ref{fig:real_robot_setup}). We collect two teleoperated demonstrations at each of five placements per task for WAM policy SFT, yielding ten demonstrations per task and 40 overall. We then run each task's SFT policy 16 times per placement, yielding 80 interaction trajectories per task and 320 overall. The Unaligned and PDA world models for a task use the same 80 trajectories. Appendix~\ref{app:reward-blind} specifies the training, calibration, and test splits for its reward classifier.

\paragraph{Execution protocol.}
All three policies use the same five predefined placements within each task. An operator restores the objects and resets the arms before each trial; the policy then acts autonomously. Ten trials per policy and placement yield 50 trials per policy and task, or 600 physical executions across the three policies and four tasks. No emergency stops or equipment failures occurred.

\begin{table*}[t]
    \centering
    \caption{\textbf{Success criteria for real-robot evaluation.}
    The final column gives examples of unsuccessful trials that reach
    an intermediate task state.}
    \label{tab:real_robot_success_criteria}
    \begin{tabular}{@{}p{0.15\textwidth}p{0.42\textwidth}p{0.37\textwidth}@{}}
        \toprule
        \textbf{Task} & \textbf{Success criterion}
        & \textbf{Example failure} \\
        \midrule
        Bottle righting
        & The robot rights a bottle lying on its side. After the gripper
          withdraws, the bottle remains upright on its base.
        & The bottle stands briefly while supported by the gripper but
          falls when the gripper withdraws. \\[3pt]
        Pen grasping
        & The robot picks up the pen from the table and places it in the
          designated container. The pen remains inside after the gripper
          releases it and withdraws.
        & The robot lifts the pen but drops it during transport, or the
          pen lands outside the container after release. \\[3pt]
        Cup stacking
        & Two cups form the base of a three-cup pyramid and jointly
          support the top cup. The pyramid remains stable after the
          gripper withdraws.
        & The top cup leans against only one base cup or slips off
          after release. \\[3pt]
        Test-tube insertion
        & The robot inserts the test tube into the designated slot in
          the rack. The tube remains in that slot after release.
        & The tube tip only touches the slot opening, the tube rests
          against the rack, or the tube enters a different slot. \\
        \bottomrule
    \end{tabular}
\end{table*}

The criteria in Table~\ref{tab:real_robot_success_criteria} require the completed task state. Each physical execution contributes one binary outcome. Task success is the total across the five placements divided by 50, and the overall rate gives equal weight to all four tasks.

\begin{table*}[t]
    \centering
    \caption{\textbf{Real-robot policy success by task.}
    Each task entry gives successes out of 50 trials and the corresponding
    success rate. The final row gives the equal-weight mean
    of the four task success rates.}
    \label{tab:real_robot_success_rates}
    \begin{tabular*}{\textwidth}{@{\extracolsep{\fill}}lccc@{}}
        \toprule
        \textbf{Task} & \textbf{SFT} & \textbf{Unaligned RL}
        & \textbf{Full PDA RL} \\
        \midrule
        Bottle righting
        & 17/50 (34\%) & 21/50 (42\%) & \textbf{25/50 (50\%)} \\
        Pen grasping
        & 13/50 (26\%) & 20/50 (40\%) & \textbf{30/50 (60\%)} \\
        Cup stacking
        & 17/50 (34\%) & 19/50 (38\%) & \textbf{24/50 (48\%)} \\
        Test-tube insertion
        & 15/50 (30\%) & 18/50 (36\%) & \textbf{23/50 (46\%)} \\
        \midrule
        Mean success rate
        & 31.0\% & 39.0\% & \textbf{51.0\%} \\
        \bottomrule
    \end{tabular*}
\end{table*}

Full PDA RL achieves the highest observed success rate on each task (Table~\ref{tab:real_robot_success_rates}). Mean success reaches 51.0\%, a gain of 20 percentage points over SFT (31.0\%) and 12 points over Unaligned RL (39.0\%). The largest improvement is on pen grasping: Full PDA RL succeeds in 30 of 50 trials, compared with 20 for Unaligned RL. A Cochran--Mantel--Haenszel test stratified by task compares Full PDA RL with Unaligned RL (102/200 vs.\ 78/200 successes, $p=0.016$) and with SFT (102/200 vs.\ 62/200, $p<10^{-4}$).

\begin{figure*}[t]
    \centering
    \includegraphics[width=\textwidth]{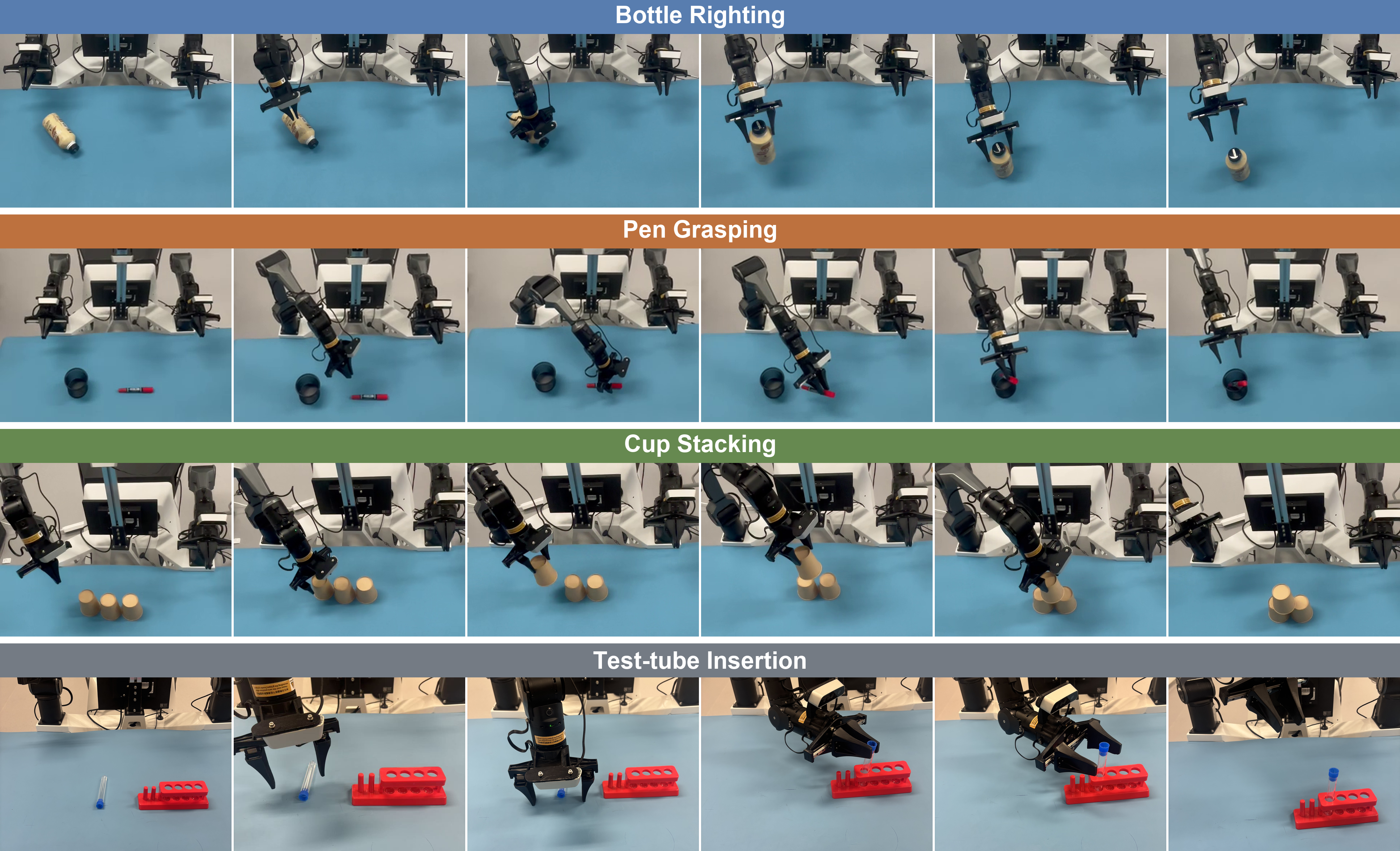}
    \caption{\textbf{Representative executions of the four real-robot tasks.}
    Frames progress from left to right. From top to bottom, the robot
    rights a bottle and releases it upright, picks up a pen and places
    it in the designated container, stacks three cups into a stable
    pyramid, and inserts a test tube into the designated rack slot.}
    \label{fig:real_robot_task_sequences}
\end{figure*}

Figure~\ref{fig:real_robot_task_sequences} shows how the completion criteria appear in practice, from releasing the upright bottle to seating the test tube in its designated slot. The trial counts establish that the mean improvement in main-paper Fig.~7 extends across all four tasks when evaluated by actual task completion at the five predefined placements.

{
    \small
    \bibliographystyle{ieeenat_fullname}
    \bibliography{main}
}

\end{document}